\documentclass[final]{cvpr}

\usepackage{times}
\usepackage{epsfig}
\usepackage{graphicx}
\usepackage{amsmath}
\usepackage{amssymb}
\usepackage{bm}
\usepackage{fixltx2e}
\usepackage{placeins}
\usepackage{gensymb}
\usepackage{tablefootnote}
\usepackage{xcolor}
\usepackage[multiple]{footmisc}
\usepackage[font=footnotesize]{caption}
\usepackage[pagebackref=true,breaklinks=true,letterpaper=true,colorlinks,bookmarks=false]{hyperref}
\usepackage[colorlinks]{hyperref}

\begin{document}

\title{Driver Glance Classification In-the-wild: Towards Generalization Across Domains and Subjects\thanks{Work done while at Smart Eye}}

\author{\parbox{16cm}{\centering
    {\large Sandipan Banerjee$^1$, Ajjen Joshi$^1$, Jay Turcot$^1$, Bryan Reimer$^2$, and Taniya Mishra*$^3$}\\
    {\normalsize
    $^1$ Smart Eye, $^2$ Massachusetts Institute of Technology, $^3$ SureStart\\
        \tt\small \{firstname.lastname\}@smarteye.ai, \tt\small reimer@mit.edu, \tt\small taniya.mishra@mysurestart.com
    }}
}

\maketitle
\vspace{-1cm}
\begin{abstract}
\vspace{-0.3cm}
Distracted drivers are dangerous drivers. Equipping advanced driver assistance systems (ADAS) with the ability to detect driver distraction can help prevent accidents and improve driver safety. In order to detect driver distraction, an ADAS must be able to monitor their visual attention. We propose a model that takes as input a patch of the driver's face along with a crop of the eye-region and classifies their glance into 6 coarse regions-of-interest (ROIs) in the vehicle. We demonstrate that an hourglass network, trained with an additional reconstruction loss, allows the model to learn stronger contextual feature representations than a traditional encoder-only classification module. To make the system robust to subject-specific variations in appearance and behavior, we design a personalized hourglass model tuned with an auxiliary input representing the driver's baseline glance behavior. Finally, we present a weakly supervised multi-domain training regimen that enables the hourglass to jointly learn representations from different domains (varying in camera type, angle), utilizing unlabeled samples and thereby reducing annotation cost.
\end{abstract}

\vspace{-0.2cm}
\section{Introduction}
\vspace{-0.2cm}
Driver distraction has been shown to be a leading cause of vehicular accidents \cite{fitch2013impact}. Anything that competes for a driver's attention, such as talking or texting on the phone, using the car's navigation system or eating, can be a cause of distraction. A distracted individual often directs their visual attention away from driving, which has been shown to increase accident risk \cite{liang2012dangerous}. According to the NHTSA, a large percentage of crashes and near-crashes occur when the driver looks away from the street \cite{klauer2006impact}.Therefore, driver glance behavior can be an important signal in determining their level of distraction. A system that can accurately detect where the driver is looking can then be used to alert drivers when their attention shifts away from the road. Such systems can also monitor driver attention to manage and motivate improved awareness \cite{coughlin2011monitoring}. For example, the system can decide whether a driver's attention needs to be cued back to the road prior to safely handing them back the control.

A real-time system that can classify driver attention into a set of ROIs can be used to infer their overall attentiveness and offer predictive indication of attention failures associated with crashes and near-crashes \cite{seppelt2017glass}. Real-time tracking of driver gaze from video is attractive because of the low equipment cost but challenging due to variations in illumination, eye occlusions caused by eyeglasses/sunglasses and poor video quality due to vehicular movements and sensor noise. In this paper, we propose a model that can predict driver glance ROI, given a patch of the driver's face along with a crop of their eye-region. We show that an hourglass network \cite{UNet,StackedHourGlass}, composed of encoder-decoder modules, trained with a reconstruction loss on top of the classification task, performs better than a vanilla CNN. The reconstruction task serves as a regularizer \cite{DecoderTTIC}, helping the model learn robust representations of the input by implicitly leveraging useful information around its context \cite{Context3,PJP_FG}. 

However, a model that makes predictions based on only a single static frame may struggle to deal with variations in subject characteristics not well represented in the training set (\emph{e.g.} a shorter or taller-than average driver may have different appearances for the default on-the-road driving behavior). To address this challenge, we add an auxiliary input stream representing the subject's baseline glance behavior, yielding improved performance over a rigid network. 

\begin{figure*}
  \centering
  \includegraphics[width=0.95\textwidth]{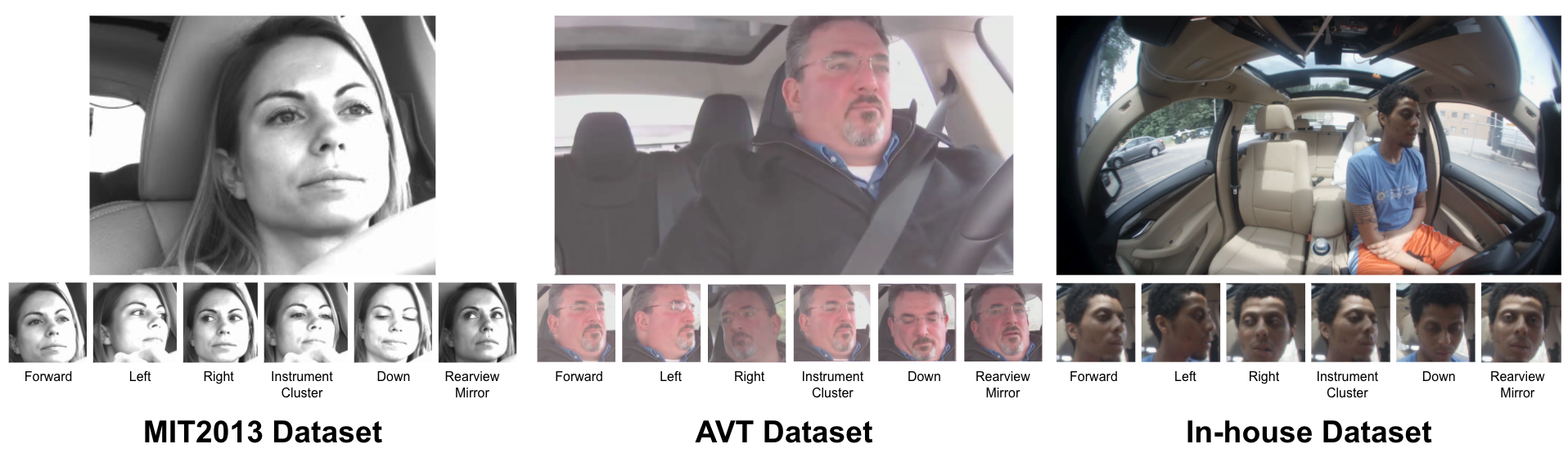}
  \caption{Sample frames from the datasets used in our experiments (MIT2013 (left), AVT (middle) and In-house (right). For each dataset, we present an example raw frame captured by the camera, and an example each of a driver’s cropped face for each driver glance region-of-interest class.}
  \label{fig:Schema}
\vspace{-0.7cm}
\end{figure*}

Another challenge associated with an end-to-end glance classification system is the variation in camera type (RGB/NIR) and placement (on the steering wheel or rearview mirror). Due to variations in cabin configuration, it is impossible to place the camera in the same location with a consistent view of the car interior and the driver. Therefore, a model trained on driver head-poses associated with a specific camera-view may not generalize. To overcome this domain-mismatch challenge, we present a framework to jointly train models in the presence of data from multiple domains (camera types and views). Leveraging our backbone hourglass' reconstruction objective, this framework can utilize unlabeled samples from multiple domains along with weak supervision to jointly learn stronger domain-invariant representations while effectively reducing labeling cost.

In summary, we make the following contributions: (1) we propose an hourglass architecture that can predict driver glance ROI from static images, illustrating the utility of adding a reconstruction loss to learn robust representations even for classification tasks; (2) we design a personalized version of our hourglass model, that additionally learns residuals in feature space from the driver's default `eyes-on-the-road' behavior, to better tune output mappings wrt the subject's default; (3) we formulate a weakly supervised multi-domain training approach that utilizes unlabeled samples for classification and allows for model adaptation to novel camera types and angles, while reducing the associated labeling cost.

\vspace{-0.4cm}
\section{Related Work}
\vspace{-0.15cm}
Computer-vision based driver monitoring systems \cite{chhabra2017survey} have been used to estimate a driver's state of fatigue \cite{joshi2020inthewild}, cognitive load \cite{fridman2018cognitive} or the driver's focus on the road \cite{vicente2015driver}. 

{\bf Gaze estimation}: The problem of tracking gaze from video has been studied extensively \cite{hansen2009eye, cazzato2020look}. Professional gaze tracking systems do exist (\emph{e.g.} Tobii\cite{tobii}), however they typically require user or session-specific calibration to achieve good performance. Appearance-based, calibration-free gaze estimation has numerous applications in computer vision, from gaze-based human-computer interaction to analysis of visual behavior. Researchers have utilized both real \cite{zhang2017mpiigaze} and synthetic data \cite{wood2015rendering, wood2016learning} to model gaze behavior, with generative approaches used to bridge the gap between synthetic and real distributions, so that models trained on one domain work well on another \cite{shrivastava2017learning, kim2019nvgaze}.

{\bf Glance Classification}: In the case of driver distraction, classifying where the driver is looking from an estimated gaze vector involves finding the intersection between the gaze vector and the 3D car geometry. A simpler alternative is to directly classify the driver image into a set of car ROIs using head pose \cite{jha2018probabilistic}, as well as eye region appearance\cite{fridman2016owl}. Rangesh et al. focused on estimating driver gaze in the presence of eye-occluding glasses to synthetically remove eyeglasses from input images before feeding them to a classification network \cite{rangesh2020driver}. Ghosh et al. recently introduced the Driver Gaze in the Wild (DGW) dataset to further encourage research in this area \cite{dgw}.

{\bf Personalization}: Personalized training has been applied to other domains (e.g. facial action unit \cite{chu2013selective} and gesture recognition \cite{yao2014gesture, joshi2017personalizing}) but not yet on vehicular glance classification. In the context of eye tracking, personalization is usually achieved through apriori user calibration. \cite{krafka2016eye} reported results for unconstrained (calibration-free) eye tracking from mobile devices and showed calibration to significantly improve performance. For personalizing gaze models latent representation for each eye has been used \cite{linden2019learning}, for utilizing saliency information in visual content \cite{chang2019salgaze} or adapting a generic example with few training samples \cite{yu2019improving}.

{\bf Domain Invariance}: Domain adaptation has been used in a variety of applications, \emph{e.g.} object recognition \cite{DA2010}. Researchers have trained shared networks with samples from different domains, regularized via an adaptation loss between their embeddings \cite{FineGrainedDom,AdvDiscDA}, or trained models with domain confusion to learn domain-agnostic embeddings \cite{DomainConfusion}, implemented by reducing distance between the domain embeddings \cite{SimDeepTrans,Peng2019DomainAL} or reversing the gradient specific to domain classification during backpropagation \cite{GradRev}. Another popular approach towards domain adaptation is to selectively fine-tune task specific models from pre-trained weights \cite{TransferLearning,ImgNetTransfer} by freezing pre-trained weights that are tuned to specific tasks or domains \cite{PiggyBack,PackNet} or selectively pruning weights \cite{PruningGuide} to prudently adapt to new domains \cite{NetTailor}. Specific to head pose, lighting and expression agnostic face recognition, approaches like feature normalization using class centers \cite{CenterLoss,RingLoss} and class separation using angular margins \cite{SphereFace,CosFace,ArcFace} have been proposed. Such recognition tasks have also benefitted from mixing samples from different domains, like real and synthetic \cite{MasiAug,SREFI2}.

\begin{figure*}
\centering
   \includegraphics[width=0.85\linewidth]{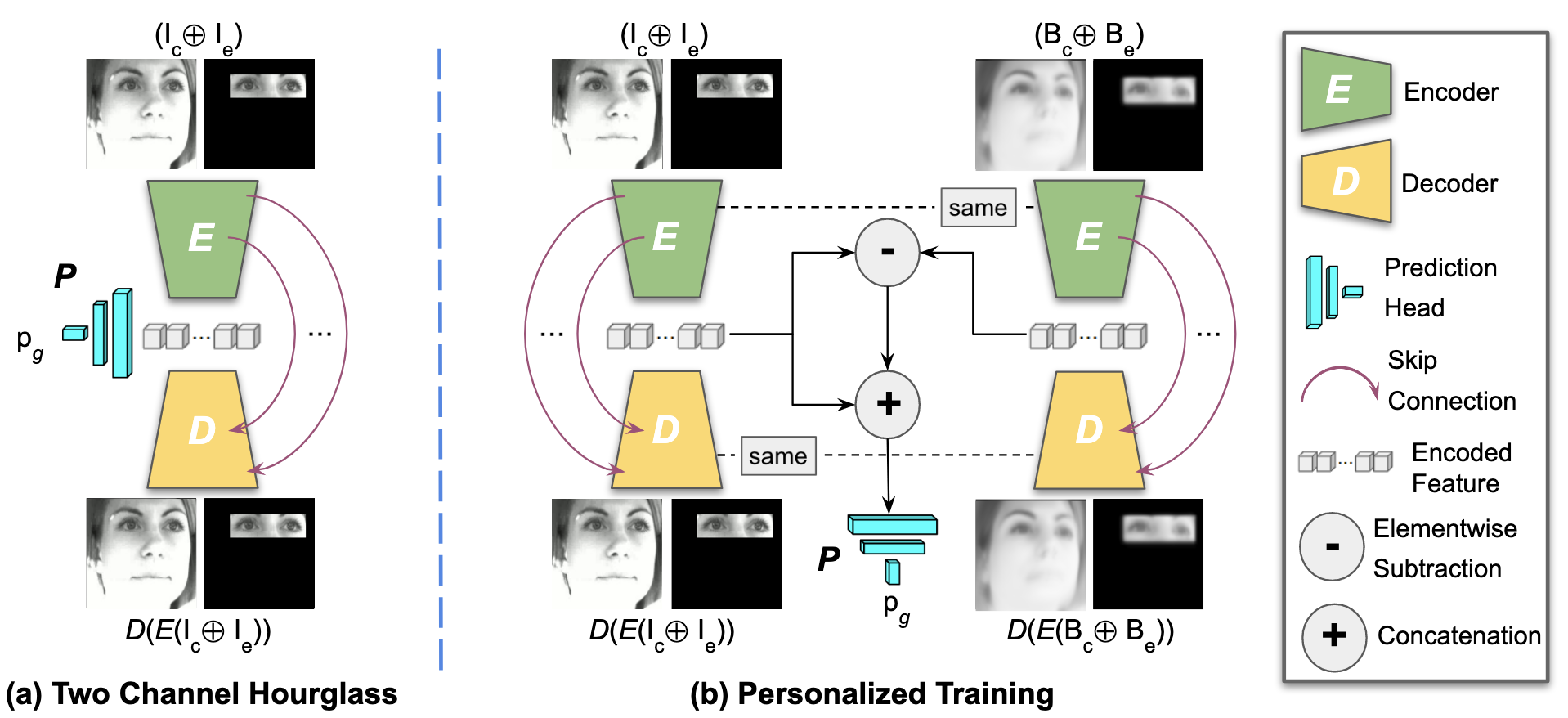}
   \caption{Illustration of our (a) two channel hourglass and (b) multi-stream personalization models described in Sections \ref{sec:2cHG} and \ref{sec:personalized} respectively.}
\label{fig:Pipeline_1}
\vspace{-0.7cm}
\end{figure*}

While most research on gaze estimation proposes models that predict gaze vectors, our glance classification model directly predicts the actual ROI of the driver's gaze inside the vehicle. Unlike previous work, our multi-domain training approach tunes the model's ROI prediction to jointly work on multiple domains (\emph{e.g.} car interiors), varying in camera type, angle and lighting, while requiring very little labeled data. Our model can be personalized for continual tuning based on the driver's behavior and anatomy as well.

\vspace{-0.2cm}
\section{Dataset Description and Data Analysis}
\vspace{-0.2cm}
{\bf  MIT-2013}: The dataset was extracted from a corpus of driver-facing videos, which were collected as part of large driving study that took place on a local interstate highway \cite{mehler2016multi}. For each participant in the study, videos of the drivers were collected either in a 2013 Chevrolet Equinox or a Volvo XC60. The participants performed a number of tasks, such as using the voice interface to enter addresses or combining it with manual controls to select phone numbers, while driving. Frames with the frontal face of the drivers were then annotated to the following ROIs: `road', `center stack', `instrument cluster', `rearview mirror', `left', `right', `left blindspot', `right blindspot', `passenger', `uncodable', and `other'. The data of interest was independently coded by two evaluators and mediated according to standards described by \cite{smith2005methodology}.  Following Fridman et al. \cite{fridman2016owl}, frames labeled `left' and `left blindspot' were given a single generic `left' label and frames labeled `right', `right blindspot' and `passenger' were given a generic 'right' label, while frames labeled `uncodable', and `other' were ignored. We used a subset of the data with 97 unique subjects, which was split into 60 train, 17 validation and 20 test subjects. 

{\bf AVT}: This dataset contains driver-initiated, non-critical disengagement events of Tesla Autopilot in naturalistic driving \cite{morando2020driver} and was extracted from a large corpus of naturalistic driving data, collected from an instrumented fleet of 29 vehicles, each of which record the IMU, GPS, CAN messages, and video streams of the driver face, the vehicle cabin and the forward roadway \cite{fridman2019advanced}. The MIT Advanced Vehicle Technology (MIT-AVT) study was designed to collect large-scale naturalistic driving data for better understanding of how drivers interact with modern cars to aid better design and interfaces as vehicles transition into increasingly automated systems. Each video was processed by a single coder with inter-rater reliability assessments as detailed in \cite{morando2020driver}.

{\bf In-house}: This dataset was collected to train machine learning models to estimate gaze from the RGB and NIR camera types and a challenging camera angle. A camera, with a wide-angle lens, was placed under the rear-view mirror for this collection, the focus of which was to capture data from a position where the entire cabin was visible. Participants followed instructions from a protocol inside a static/parked car, where they glanced at various ROIs using 3 behavior types: `owl', `lizard' and `natural'\cite{fridman2016owl}. In our experiments, we used samples from 85 participants - 50 for training, 18 for validation and 17 for testing. Videos of each participant was manually annotated by 3 human labelers. Example frames from all three datasets are shown in Figure \ref{fig:Schema}.

We do not use the recently released DGW dataset \cite{dgw} in our experiments as the annotations are provided for a different set of regions, with the driver seated on the right-hand side, which makes it difficult to integrate into our multi-domain training pipeline.

\vspace{-0.2cm}
\section{Proposed Models}\label{sec:models}
\vspace{-0.1cm}
\subsection{Two-channel Hourglass}\label{sec:2cHG}
\vspace{-0.1cm}
While a standalone classification (\emph{i.e.} encoder with prediction head) or reconstruction module (\emph{i.e.} encoder-decoder) can produce high performance numbers for recognition or semantic segmentation or super-resolution tasks, combining them together has been shown to further boost model performance \cite{FB_Reg,AdvAuto,DecoderTTIC,HarisTTIC,SharmaTask}. The auxiliary module's (prediction or reconstruction) loss acts as a regularizer \cite{DecoderTTIC} and boosts model performance on the primary task. For our specific task of driver glance classification, adding a reconstruction element can tune the model weights to implicitly pay close attention to contextual pixels while making a decision. Thus, instead of using a feed forward neural network, as traditionally done for classification tasks \cite{AlexNet,vgg_sim,ResNet,ILSVRC15}, we use an hourglass structure consisting of encoder ($E$) and decoder ($D$) modules \cite{UNet}. 

\begin{figure*}
\centering
   \includegraphics[width=0.8\linewidth]{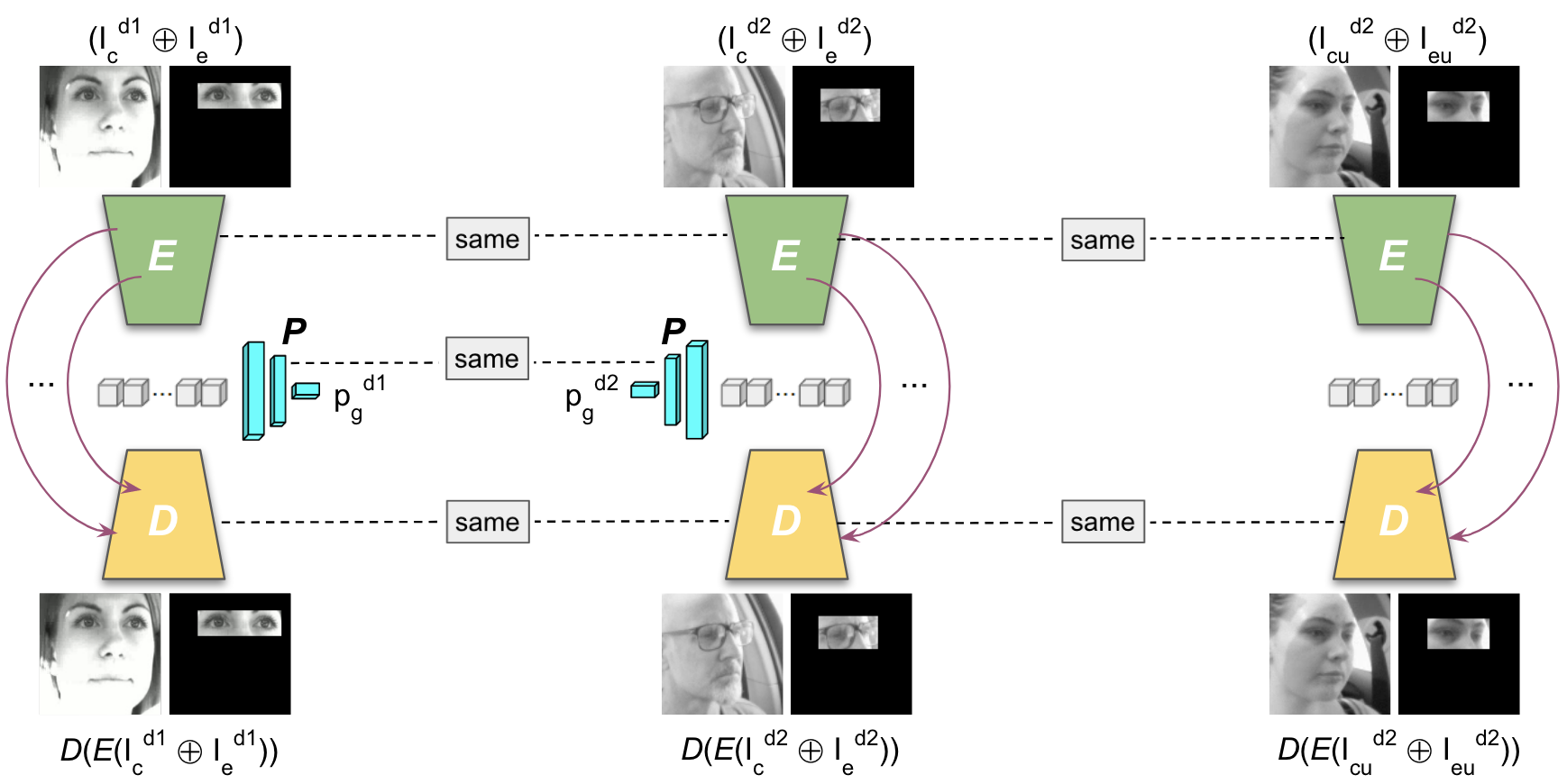}
   \caption{Our multi-domain training pipeline: For every iteration, the model is trained with mini-batches consisting of labeled input samples from $d1$ ($(\text{I}^{d1}_{c} \oplus \text{I}^{d1}_{e})$) and $d2$ ($(\text{I}^{d2}_{c} \oplus \text{I}^{d2}_{e})$), and unlabeled input from $d2$ ($(\text{I}^{d2}_{cu} \oplus \text{I}^{d2}_{eu})$). The model weights are updated based on the overall loss accumulated over the mini-batches. It is to be noted that all three subjects in this figure are looking at the road, but appear very different due to different camera angles.}
\label{fig:Pipeline_2}
\vspace{-0.75cm}
\end{figure*}

In our model, $E$ takes as input the cropped face and eye patch images $\text{I}_c$ and $\text{I}_e$ respectively, concatenated together as a two-channel tensor $(\text{I}_c \oplus \text{I}_e)$ and produces a feature vector (\emph{i.e.} $E(\text{I}_c \oplus \text{I}_e)$) as its encoded representation. This feature vector is then passed through a prediction head $P$ to extract the estimated glance vector {\bf $\text{p}_g$}, before being sent to $D$ to generate the face and eye patch reconstructions $D(E(\text{I}_c \oplus \text{I}_e))$, as shown in Figure \ref{fig:Pipeline_1}.a. $E$ is composed of a dilated convolution layer \cite{atrous} followed by a set of n downsampling residual blocks \cite{ResNet} and a dense layer for encoding. $D$ takes $E(\text{I}_c \oplus \text{I}_e)$ and passes it through n upsampling pixel shuffling blocks \cite{pixshuff} followed by a convolution layer with \emph{tanh} activation for image reconstruction \cite{DCGAN,salimans}. For better signal propagation, we add skip connections \cite{UNet} between corresponding layers in $E$ and $D$ \cite{SREFI3}. The encoded feature is also passed through the prediction head $P$, composed of two densely connected layers followed by softmax activation to produce the glance prediction vector {\bf $\text{p}_g$}. 

The hourglass model is trained using a categorical cross entropy based classification loss $L_{cls}$ between the ground truth glance vector {\bf $\text{c}_g$} and the predicted glance vector $P(E(\text{I}_c \oplus \text{I}_e))$ (\emph{i.e.} {\bf $\text{p}_g$}), and a pixelwise reconstruction loss $L_{rec}$ between the input tensor $(\text{I}_c \oplus \text{I}_e)$ and its reconstruction $D(E(\text{I}_c \oplus \text{I}_e))$. For a given training batch $\mathcal{N}$ and the ground truth classes $\mathcal{C}$, they can be represented as:
\vspace{-0.2cm}
\begin{equation}
L_{cls} = -\frac{1}{\left | \mathcal{N} \right |}\sum_{1}^{\mathcal{N}}\sum_{i}^{\mathcal{C}}\text{c}_{gi}\log P(E(\text{I}_c \oplus \text{I}_e))_{i}
\label{eq:Lcls}
\vspace{-0.2cm}
\end{equation}
\begin{equation}
L_{rec} = \frac{1}{\left | \mathcal{N} \right |}\sum_{1}^{\mathcal{N}}\left | (\text{I}_c \oplus \text{I}_e) - D(E(\text{I}_c \oplus \text{I}_e)) \right |
\label{eq:Lrec}
\end{equation}
where $\mathcal{N}$ is the training set in a batch and $\mathcal{C}$ the ground truth classes. The overall objective $L$ is defined as:
\begin{equation}
L = L_{cls} + \lambda_1L_{rec}
\label{eq:L}
\vspace{-0.1cm}
\end{equation}

\subsection{Personalized Training}\label{sec:personalized}
\vspace{-0.2cm}
As mentioned earlier, introducing an auxiliary channel of baseline information can better tune the classification model to specific driver anatomy and behaviors. To this end, we propose a personalized version of our hourglass framework, composed of the same encoder ($E$) and decoder ($D$) modules. For each driver in the training dataset, we extract their baseline face crop $\text{B}_e$ and eye patch $\text{B}_e$, where $\text{B}_e = \frac{1}{m}\sum_{i=1}^{m}\text{I}_e$ and $\text{B}_c = \frac{1}{m}\sum_{i=1}^{m}\text{I}_c$, for all cases where the driver is looking forward at the road. The baseline face crop and eye patch images are calculated prior to training. 

During training, we extract the representation of the current frame $E(\text{I}_c \oplus \text{I}_e)$ by passing the face crop $\text{I}_c$ and eye patch $\text{I}_e$ images through $E$. Additionally, the baseline representation of the driver $E(\text{B}_c \oplus \text{B}_e)$ is computed by utilizing the baseline images. The residual between these tensors is computed in the representation space using encoded features as $E(\text{I}_c \oplus \text{I}_e) -  E(\text{B}_c \oplus \text{B}_e)$. This residual acts as a measure of variance of the driver's glance behavior from looking forward, and is concatenated with the current frame representation $E(\text{I}_c \oplus \text{I}_e)$. This concatenated tensor is then passed through the prediction head $P$ to get the glance prediction {\bf $\text{p}_g$}. Two streams each for $E$ and $D$ are deployed during training that share weights, as depicted in Figure \ref{fig:Pipeline_1}.b.

The classification loss $L^{p}_{cls}$ is then calculated as:
\vspace{-0.3cm}
\begin{multline}
L^{p}_{cls} = -\frac{1}{\left | \mathcal{N} \right |}\sum_{1}^{\mathcal{N}}\sum_{i}^{\mathcal{C}}\text{c}_{gi}\log P((E(\text{I}_c \oplus \text{I}_e) - \\ E(\text{B}_c \oplus \text{B}_e)) \oplus E(\text{I}_c \oplus \text{I}_e))_{i}
\label{eq:Lcls_p}
\end{multline}
where $\mathcal{N}$ is training batch and $\mathcal{C}$ the ground truth classes. 

The reconstruction loss $L^{p}_{rec}$ is calculated for both the current frame and baseline tensors as:
\vspace{-0.2cm}
\begin{multline}
L^{p}_{rec} = \frac{1}{\left | \mathcal{N} \right |}\sum_{1}^{\mathcal{N}}\left | (\text{I}_c \oplus \text{I}_e) - D(E(\text{I}_c \oplus \text{I}_e)) \right | + \\ \frac{1}{\left | \mathcal{N} \right |}\sum_{1}^{\mathcal{N}}\left | (\text{B}_c \oplus \text{B}_e) - D(E(\text{B}_c \oplus \text{B}_e)) \right |
\label{eq:Lrec_p}
\end{multline}
The overall objective $L^{p}$ is a weighted sum of these two losses, calculated as:
\vspace{-0.1cm}
\begin{equation}
L^{p} = L^{p}_{cls} + \lambda_2L^{p}_{rec}
\label{eq:L_p}
\vspace{-0.1cm}
\end{equation}

\subsection{Domain Invariance}\label{sec:domain}
\vspace{-0.2cm}
As can be seen in Figure \ref{fig:Schema}, driver glance can look significantly different when the camera type (RGB or NIR), its placement (steering wheel or rear-view mirror) and car interior changes. Such a domain mismatch can result in considerable decrease in performance when the classification model is trained on one dataset and tested on another, as experimentally shown in Section \ref{sec:Experiments}. To mitigate this domain inconsistency problem, we propose a multi-domain training regime for our two-channel hourglass model. This regime leverages a rich set of labeled training images from one domain to learn domain invariant features for glance estimation from training samples from a second domain, only some of which are labeled. The hourglass structure of our model provides an advantage as the unlabeled samples from the second domain can also be utilized during training using $D$'s reconstruction error.

Our multi-domain training starts with three input tensors:\\
(1) $(\text{I}^{d1}_{c} \oplus \text{I}^{d1}_{e})$ - the labeled face crop and eye patch images from the richly labeled domain $d1$,\\
(2) $(\text{I}^{d2}_{c} \oplus \text{I}^{d2}_{e})$ - the labeled face crop and eye patch images from the sparsely labeled second domain $d2$,\\
(3) $(\text{I}^{d2}_{cu} \oplus \text{I}^{d2}_{eu})$ - the unlabeled face crop and eye patch images from the second domain $d2$. 

Each tensor is passed through $E$ to generate their embedding, which are then passed through $D$ to reconstruct the input. For the input tensors with glance labels (\emph{i.e.} $(\text{I}^{d1}_{c} \oplus \text{I}^{d1}_{e})$ and $(\text{I}^{d2}_{c} \oplus \text{I}^{d2}_{e})$), the encoded feature is also passed through $P$ to get the glance predictions {\bf $\text{p}^{d1}_g$} and {\bf $\text{p}^{d2}_g$} respectively. We set shareable weights across the multi-streams of $E$, $D$ and $P$ during training, as shown in Figure \ref{fig:Pipeline_2}. 

The classification loss $L^{md}_{cls}$ for the multi-domain training is set as:
\vspace{-0.3cm}
\begin{multline}
L^{md}_{cls} = -\frac{1}{\left | \mathcal{N}^{d1} \right |}\sum_{1}^{\mathcal{N}^{d1}}\sum_{i}^{\mathcal{C}}\text{c}^{d1}_{gi}\log P(E(\text{I}^{d1}_c \oplus \text{I}^{d1}_e))_{i} - \\ \frac{1}{\left | \mathcal{N}^{d2} \right |}\sum_{1}^{\mathcal{N}^{d2}}\sum_{i}^{\mathcal{C}}\text{c}^{d2}_{gi}\log P(E(\text{I}^{d2}_c \oplus \text{I}^{d2}_e))_{i}
\label{eq:Lcls_md}
\end{multline}
where $\mathcal{N}^{d1}$ and $\mathcal{N}^{d2}$ are the labeled training batches, and $\text{c}^{d1}_{g}$ and $\text{c}^{d2}_{g}$ are the ground truth glance labels from domains $d1$ and $d2$ respectively. 

Similarly, the reconstruction error $L^{md}_{rec}$ is calculated as:
\vspace{-0.2cm}
\begin{multline}
L^{md}_{rec} = \frac{1}{\left | \mathcal{N}^{d1} \right |}\sum_{1}^{\mathcal{N}^{d1}}\left | (\text{I}^{d1}_c \oplus \text{I}^{d1}_e) - D(E(\text{I}^{d1}_c \oplus \text{I}^{d1}_e)) \right | + \\ \frac{1}{\left | \mathcal{N}^{d2} \right |}\sum_{1}^{\mathcal{N}^{d2}}\left | (\text{I}^{d2}_c \oplus \text{I}^{d2}_e) - D(E(\text{I}^{d2}_c \oplus \text{I}^{d2}_e)) \right | + \\ \frac{1}{\left | \mathcal{N}^{d2}_{u} \right |}\sum_{1}^{\mathcal{N}^{d2}_{u}}\left | (\text{I}^{d2}_{cu} \oplus \text{I}^{d2}_{eu}) - D(E(\text{I}^{d2}_{cu} \oplus \text{I}^{d2}_{eu})) \right |
\label{eq:Lrec_md}
\end{multline}
where $\mathcal{N}^{d2}_{u}$ is the unlabeled training batch from domain $d2$.

The full multi-domain loss $L^{md}$ is calculated as:
\begin{equation}
L^{md} = L^{md}_{cls} + \lambda_3L^{md}_{rec}
\label{eq:L_md}
\end{equation}
The weighing scalars $\lambda_1$ (\ref{eq:L}), $\lambda_3$ (\ref{eq:L_p}) and $\lambda_3$ (\ref{eq:L_md}) are hyper-parameters that are tuned experimentally.

\vspace{-0.2cm}
\section{Experiments}\label{sec:Experiments}
\subsection{Training Details}\label{sec:TrainDet}
\vspace{-0.1cm}
To train our models we use $\sim$235K video frames from the MIT2013 dataset, and $\sim$153K and $\sim$163K for validation and testing respectively. The videos were split offline to assign into training, validation and testing buckets. Due to the large amount of labeled samples, we also use this dataset to represent the richly labeled domain (\emph{i.e.} $d1$) for our domain invariant experiments, while using the AVT or In-house datasets as the second domain $d2$ (check Section \ref{sec:domain}). We randomly sample $\sim$204K frames (Training: 162K, Validation: 22K, Testing: 20K) from the AVT and $\sim$377K video frames (Training: 240K, Validation: 65K, Testing: 72K) from the In-house datasets for these experiments. All frames were downsampled to 96$\times$96$\times$1 to generate the facial image and the eye patch was cropped out (also 96$\times$96$\times$1 in size) using the eye-landmarks extracted using \cite{Bulat3D}. Frames with undetected faces were removed.

During training, we use the Adam optimizer \cite{Adam} with the base learning rate set as $10^{-4}$ with a Dropout \cite{Dropout} layer (rate=0.7) between the dense layers of the prediction head in the two-channel hourglass network (Section \ref{sec:2cHG}). The weighing scalars $\lambda_1$, $\lambda_2$ and $\lambda_3$ are empirically set as 1, 1 and 10 respectively. We train all models using Tensorflow \cite{TF} coupled with Keras \cite{chollet2015keras} on a single NVIDIA Tesla V100 card with the batch size set as 8. For the personalized model however, we find it optimal to train with a batch size of 16 and learning rate of $10^{-3}$. To reduce computation cost and further prevent overfitting, we stop model training once the validation loss plateaus across three epochs and save the model snapshot for testing. To compute statistical significance between the performance of various methods, we train each model 5 times using random seeds for initialization. We only use the trained encoder and prediction head during inference.

For training the personalization framework, we prepare multiple mini-batches for every iteration with the current frame ($\text{I}_c$, $\text{I}_e$) and baseline frame ($\text{B}_c$, $\text{B}_e$) inputs. For the domain invariant regimen, the mini-batches are prepared with labeled $d1$ ($(\text{I}^{d1}_{c} \oplus \text{I}^{d1}_{e})$), labeled $d2$ ($(\text{I}^{d2}_{c} \oplus \text{I}^{d2}_{e})$) and unlabeled $d2$ inputs ($(\text{I}^{d2}_{cu} \oplus \text{I}^{d2}_{eu})$). The overall loss is computed from the mini-batches before updating model weights.

{\bf Computation Overhead}: In terms of model size, the encoder $E$ and prediction head $P$ together consist of 24M parameters while adding the decoder $D$ for reconstruction increases the number to 54M. While $D$ does add computational load during training, only $E$ and $P$ together are required for inference. Thus, turning the typical classifier into an hourglass does not introduce additional overhead when deployed in production. The personalized version of the model has the same number of trainable parameters but does require an additional stream of baseline driver information.

\vspace{-0.1cm}
\subsection{Performance on the MIT2013 Dataset}
\vspace{-0.1cm}
Post training, we test our two-channel hourglass and personalization models for glance estimation on the test frames from the MIT2013 dataset\cite{mehler2016multi}. To gauge of their effectiveness, we compare our model with the following:\\
(1) {\bf Landmarks + MLP}. Following \cite{fridman2016owl}, we train a baseline MLP model with 3 dense layers on a flattened representation of facial landmarks extracted using \cite{Bulat3D}.

\noindent (2) {\bf Dense Eye-landmarks + Headpose}. This recently-proposed lightweight MLP model is trained on a set of dense landmarks of the eyes, as well as head pose estimates \cite{dari_gaze}. 

\noindent (3) {\bf Baseline CNN}. We also train a baseline CNN with 4 convolutional and max pooling layers followed by 3 dense layers, similar to AlexNet \cite{AlexNet}. The baseline CNN takes as input the 96$\times$96$\times$1 cropped face image.

\noindent (4) {\bf Upperface Squeezenet}. Following the best performing configuration in \cite{vora2018driver}, we train a SqueezeNet model \cite{squeezenet} on the upper half of the driver's face. 

\noindent (5) {\bf One-Channel Hourglass}. This model only receives the cropped face image $\text{I}_{c}$ without the eye-patch channel $\text{I}_{e}$. The hyper-parameters and losses however remain the same. 

\noindent (6) {\bf Upperface Squeezenet w/ decoder}. To test whether a secondary reconstruction task helps the encoder learn more discriminable representations, we also add a decoder (same architecture as (3)), to the SqueezeNet configuration (3).

\begin{table}
\begin{center}
\captionsetup{justification=centering}
\caption{Performance (ROC-AUC) of the different glance classification models on the MIT2013 dataset. The best two results are highlighted.}
\begin{small}
\begin{tabular}{  | c| c| c| c| c| c| c| c| }
\hline
\begin{tabular}[x]{@{}c@{}}{\bf Model}\end{tabular} & 
\begin{tabular}[x]{@{}c@{}}{\bf Macro}\\{\bf Average}\end{tabular}\\
\hline
\hline
  \begin{tabular}[x]{@{}c@{}}Landmarks + MLP \cite{fridman2016owl}\end{tabular} & 0.898 $\pm$ 0.001 \\
  \hline
  \begin{tabular}[x]{@{}c@{}} Dense Eye Landmarks + Headpose \cite{dari_gaze}\end{tabular} & 0.800 $\pm$ 0.020 \\
  \hline
  \begin{tabular}[x]{@{}c@{}}Baseline CNN \cite{AlexNet}\end{tabular} & 0.953 $\pm$ 0.001 \\
  \hline
    
  \begin{tabular}[x]{@{}c@{}} Upperface SqueezeNet \cite{vora2018driver}\end{tabular} & 0.960 $\pm$ 0.002 \\
  \hline
  
  \begin{tabular}[x]{@{}c@{}}One-Channel Hourglass\end{tabular} & 0.961 $\pm$ 0.001 \\
  \hline
  \begin{tabular}[x]{@{}c@{}}Upperface SqueezeNet \cite{vora2018driver} w/ Decoder \end{tabular} & 0.964 $\pm$ 0.001 \\
  \hline
  \begin{tabular}[x]{@{}c@{}}Two-Channel Hourglass (ours)\end{tabular} & {\bf 0.966 $\pm$ 0.001} \\
\hline
  \begin{tabular}[x]{@{}c@{}}Personalized Hourglass (ours)\end{tabular} & {\bf 0.967 $\pm$ 0.001} \\
  \hline
\end{tabular}
\label{Tab:MIT_Res}
\end{small}
\end{center}
\vspace{-0.8cm}
\end{table}

As can be seen in Table \ref{Tab:MIT_Res}, increasing the input quality (\emph{e.g.} landmarks vs. actual pixels) and model complexity (\emph{e.g.} baseline CNN vs. residual encoder) also improves classification performance, with both the personalization multi-stream and hourglass models outperforming the other approaches and the latter producing the best macro average ROC-AUC. This suggests providing the model with an additional stream of subject-specific information (\emph{i.e.} personalization) can better tune the model with respect to movement of the driver head. Note that the model trained on dense eye landmarks performs poorly because the landmarks aren't localized accurately due to low input face resolution. Alternatively, adding an auxiliary reconstruction task can also boost the classification accuracy by learning useful contextual information while requiring no extra data stream, further underpinned by adding a decoder to the Squeezenet \cite{squeezenet} architecture from \cite{vora2018driver}. Additionally, our model also outperform two recent approaches \cite{vora2018driver,dari_gaze} on gaze region estimation.

\vspace{-0.1cm}
\subsection{Domain Invariance}
\vspace{-0.1cm}
For the domain invariance task, as described in Section \ref{sec:domain}, we assign the MIT2013 dataset as the richly labeled domain $d1$ as it has a large number of video frames with human-annotated glance labels and use the AVT and our In-house datasets interchangeably as the new domain $d2$. To evaluate its effectiveness, we compare our multi-domain training approach with following regimes while keeping the backbone network (two-channel hourglass) the same:\\
(1) {\bf Mixed Training}. Only labeled data from $d1$ and $d2$ are pooled together based on their glance labels for training.\\
(2) {\bf Fine-tuning} \cite{ImageGPT}. We train the model on labeled data from $d1$ and then fine-tune the saved snapshot on labeled data from $d2$, a strategy similar to \cite{ImageGPT}.\\
(3) {\bf Gradient Reversal} \cite{GradRev}. We add a domain classification block on top of the encoder output to predict the domain of each input. However, its gradient is reversed during backpropagation to confuse the model and shift its representations towards a common manifold, similar to \cite{GradRev} (we use the implementation from \cite{GradRevDrive}).\\
(4) {\bf Tri-training} \cite{tri_train,tri_train_ruder}. We split the labeled data from $d1$ and $d2$ into three disjoint sets and train two model instances in a supervised manner independently with the first two splits. Then we use these two trained instances to predict the glance state for the samples from the remaining set. If the two predictions are in agreement, we assign it as the proxy-label to the sample and use it to train the third model. This model is finally used for inference.\\
(5) {\bf Distillation} \cite{distillation}. We split the data from both domains into two sets and train a teacher model on the first split in a supervised fashion. While we use labels for the $d1$ split to train a student model, we use the teacher's prediction on the unused split from $d2$ to regress the student to its output distribution. The trained student is used for inference.

\begin{table}
\begin{center}
\captionsetup{justification=centering}
\caption{Multi-domain performance (ROC-AUC) of our hourglass model, trained using different regimes, on the {\color{red}MIT2013} and {\color{blue}AVT} datasets. The best two results are highlighted.}
\begin{small}
\begin{tabular}{  | c| c| c| c| c| c| c| c| }
\hline
\begin{tabular}[x]{@{}c@{}}{\bf Model}\end{tabular} &
\begin{tabular}[x]{@{}c@{}}{\bf Macro Average}\end{tabular}\\
\hline
\hline
  \begin{tabular}[x]{@{}c@{}}Mixed Training\end{tabular} & \begin{tabular}[x]{@{}c@{}}{\bf \color{red}0.963 $\pm$ 0.001}, {\color{blue}0.918 $\pm$ 0.005}\end{tabular} \\
  \hline
    \begin{tabular}[x]{@{}c@{}}Fine-tuning \cite{ImageGPT}\end{tabular} &  \begin{tabular}[x]{@{}c@{}}{\color{red}0.875 $\pm$ 0.001}, {\bf \color{blue}0.920 $\pm$ 0.001}\end{tabular} \\
  \hline
    \begin{tabular}[x]{@{}c@{}}Gradient Reversal \cite{GradRev}\end{tabular} &  \begin{tabular}[x]{@{}c@{}}{\color{red}0.961 $\pm$ 0.001}, {\color{blue}0.912 $\pm$ 0.001}\end{tabular} \\
  \hline
  \begin{tabular}[x]{@{}c@{}}Tri-training \cite{tri_train_ruder}\end{tabular} &  \begin{tabular}[x]{@{}c@{}}{\color{red}0.956 $\pm$ 0.002}, {\bf \color{blue}0.929 $\pm$ 0.004}\end{tabular} \\
  \hline
  \begin{tabular}[x]{@{}c@{}}Distillation \cite{distillation}\end{tabular} &  \begin{tabular}[x]{@{}c@{}}{\color{red}0.961 $\pm$ 0.001}, {\color{blue}0.895 $\pm$ 0.002}\end{tabular} \\
  \hline
    \begin{tabular}[x]{@{}c@{}}Ours \end{tabular} & \begin{tabular}[x]{@{}c@{}}{\bf \color{red}0.964 $\pm$ 0.001}, {\color{blue}0.919 $\pm$ 0.001}\end{tabular} \\
  \hline
\end{tabular}
\label{Tab:MIT_AVT_Res}
\end{small}
\end{center}
\vspace{-0.5cm}
\end{table}

\begin{table}
\begin{center}
\captionsetup{justification=centering}
\caption{Multi-domain performance (ROC-AUC) of our hourglass model, trained using different regimes, on the {\color{red}MIT2013} and the {\color{cyan}In-house} dataset. The best two results are highlighted.}
\begin{small}
\begin{tabular}{  | c| c| c| c| c| c| c| c| }
\hline
\begin{tabular}[x]{@{}c@{}}{\bf Model}\end{tabular} &
\begin{tabular}[x]{@{}c@{}}{\bf Macro Average}\end{tabular}\\
\hline
\hline
  \begin{tabular}[x]{@{}c@{}}Mixed Training\end{tabular} & \begin{tabular}[x]{@{}c@{}}{\bf \color{red}0.956 $\pm$ 0.002}, {\bf \color{cyan}0.882 $\pm$ 0.002}\end{tabular} \\
  \hline
    \begin{tabular}[x]{@{}c@{}}Fine-tuning \cite{ImageGPT}\end{tabular} &  \begin{tabular}[x]{@{}c@{}}{\color{red}0.807 $\pm$ 0.001}, {\color{cyan}0.858 $\pm$ 0.001}\end{tabular} \\
  \hline
    \begin{tabular}[x]{@{}c@{}}Gradient Reversal \cite{GradRev}\end{tabular} &  \begin{tabular}[x]{@{}c@{}}{\color{red}0.939 $\pm$ 0.002}, {\color{cyan}0.849 $\pm$ 0.003}\end{tabular} \\
  \hline
  \begin{tabular}[x]{@{}c@{}}Tri-training \cite{tri_train_ruder}\end{tabular} &  \begin{tabular}[x]{@{}c@{}}{\color{red}0.953 $\pm$ 0.001}, {\color{cyan}0.840 $\pm$ 0.005}\end{tabular} \\
  \hline
  \begin{tabular}[x]{@{}c@{}}Distillation \cite{distillation}\end{tabular} &  \begin{tabular}[x]{@{}c@{}}{\color{red}0.945 $\pm$ 0.001}, {\color{cyan}0.830 $\pm$ 0.001}\end{tabular} \\
  \hline
    \begin{tabular}[x]{@{}c@{}}Ours \end{tabular} & \begin{tabular}[x]{@{}c@{}}{\bf \color{red}0.962 $\pm$ 0.001}, {\bf \color{cyan}0.877 $\pm$ 0.003}\end{tabular} \\
  \hline
\end{tabular}
\label{Tab:MIT_Aff_Res}
\end{small}
\end{center}
\vspace{-1cm}
\end{table}

Although our multi-domain training approach can utilize the unlabeled samples from $d2$, for our first experiment we use 100\% of the annotated images from both $d1$ and $d2$ to level the playing field. The same model snapshot is used for testing on both the MIT2013 dataset ($d1$) and the AVT or In-house datasets ($d2$). The results can be seen in Tables \ref{Tab:MIT_AVT_Res} and \ref{Tab:MIT_Aff_Res} respectively. In both cases, the fine-tuning approach fails to generalize to both domains, essentially ``forgetting'' details of the initial task (\emph{i.e.} $d1$). Adding the gradient reversal head, does generate a boost over fine-tuning, however it overfits slightly on the training set and takes almost twice as the other approaches to converge. The tri-training and distillation based models generate competitive scores but fail to glean the full information from both domains due to the noise in the proxy labels. The mixed training and our multi-domain approaches perform competitively and generate the best two ROC-AUC numbers overall. 

\begin{table*}
\begin{center}
\captionsetup{justification=centering}
\caption{Performance (ROC-AUC) of our two-channel hourglass model with mixed training and our multi-domain regime, on the In-house dataset with different amount of labeled samples. The utility of unlabeled samples paired with reconstruction loss is evident as the percentage of labeled data from the second domain decreases.}
\begin{small}
\begin{tabular}{  | c| c| c| c| c| c| c| c| }
\hline
\begin{tabular}[x]{@{}c@{}}{\bf Model (labeled data)}\end{tabular} & \begin{tabular}[x]{@{}c@{}}{\bf Centerstack}\end{tabular} & \begin{tabular}[x]{@{}c@{}}{\bf Instrument Cluster}\end{tabular} & \begin{tabular}[x]{@{}c@{}}{\bf Left}\end{tabular} & \begin{tabular}[x]{@{}c@{}}{\bf Rearview Mirror}\end{tabular} & 
\begin{tabular}[x]{@{}c@{}}{\bf Right}\end{tabular} & 
\begin{tabular}[x]{@{}c@{}}{\bf Road}\end{tabular} & 
\begin{tabular}[x]{@{}c@{}}{\bf Macro Average}\end{tabular}\\
\hline
\hline
  \begin{tabular}[x]{@{}c@{}}Mixed Training (50\%)\end{tabular} & 0.876 & 0.785 & 0.920 & 0.936 & 0.922 & 0.828 & 0.877 \\
  \hline
  \begin{tabular}[x]{@{}c@{}}Ours (50\%)\end{tabular} & {\bf 0.897} & {\bf 0.818} & {\bf 0.944} & {\bf 0.943} & {\bf 0.938} & {\bf 0.842} & {\bf 0.897} \\
  \hline
  \hline
  \begin{tabular}[x]{@{}c@{}}Mixed Training (10\%)\end{tabular} & 0.821 & 0.734 & 0.882 & 0.867 & {\bf 0.924} & 0.798 & 0.838 \\
  \hline
  \begin{tabular}[x]{@{}c@{}}Ours (10\%)\end{tabular} & {\bf 0.881} & {\bf 0.814} & {\bf 0.900} & {\bf 0.898} & 0.893 & {\bf 0.845} & {\bf 0.872} \\
  \hline
  \hline
    \begin{tabular}[x]{@{}c@{}}Mixed Training (1\%)\end{tabular} & 0.776 & 0.704 & 0.859 & 0.775 & {\bf 0.854} & 0.696 & 0.777 \\
  \hline
  \begin{tabular}[x]{@{}c@{}}Ours (1\%)\end{tabular} & {\bf 0.830} & {\bf 0.790} & {\bf 0.904} & {\bf 0.847} & 0.852 & {\bf 0.777} & {\bf 0.833} \\
  \hline
\end{tabular}
\label{Tab:Aff_Res}
\end{small}
\end{center}
\vspace{-0.7cm}
\end{table*}

\begin{table}
\begin{center}
\captionsetup{justification=centering}
\caption{Performance (ROC-AUC) of our two-channel hourglass model with different components ablated on the MIT2013 dataset.}
\begin{small}
\begin{tabular}{  | c| c| c| c| c| c| c| c| }
\hline
\begin{tabular}[x]{@{}c@{}}{\bf Model}\end{tabular} & 
\begin{tabular}[x]{@{}c@{}}{\bf Macro Average}\end{tabular}\\
\hline
\hline
  \begin{tabular}[x]{@{}c@{}}w/ MSE\end{tabular}  & 0.963 $\pm$ 0.001 \\
  \hline
  \begin{tabular}[x]{@{}c@{}}w/o skip connections\end{tabular} & 0.961 $\pm$ 0.002 \\
  \hline
  \begin{tabular}[x]{@{}c@{}}w/o $L_{rec}$\end{tabular} & 0.959 $\pm$ 0.001 \\
  \hline
  \begin{tabular}[x]{@{}c@{}}w/o $L_{cls}$ \cite{ImageGPT}\end{tabular} & 0.962 $\pm$ 0.001 \\
  \hline
  \begin{tabular}[x]{@{}c@{}}Full Model\end{tabular} & {\bf 0.966 $\pm$ 0.001} \\
  \hline
\end{tabular}
\label{Tab:Ablation}
\end{small}
\end{center}
\vspace{-0.7cm}
\end{table}

However, using all labeled data from the new domain does not fairly evaluate the full potential of our approach. Unlike the other approaches, our training regimen can utilize the unlabeled data (\emph{i.e.} $(\text{I}^{d2}_{cu} \oplus \text{I}^{d2}_{eu})$) via the reconstruction loss, as proposed in Section \ref{sec:domain}. To put this functionality into effect, we use different amount of labeled samples (50\%, 10\% and 1\%) from $d2$ during training the hourglass model with mixed training and multi-domain regimes. As shown in Table \ref{Tab:Aff_Res}, our approach significantly outperforms mixed training as the amount of labeled data in the new domain diminishes. Interestingly, our multi-domain hourglass trained with 50\% labeled data generalizes better than when trained with 100\% labeled data suggesting more generalizable global features are learned when an unsupervised component is added to a classification task. Thus, this technique can gauge the amount of labeling required when adapting models to new domains and consequently reduce annotation cost.

\subsection{Ablation Studies}
To check the contribution of each component, we train the following variations of our hourglass model:\\
(1) {\bf w/ MSE}. Instead of mean absolute error, the reconstruction loss is computed with mean squared error.\\
(2) {\bf w/o skip connections}. We remove skip connections between the encoder and decoder layers.\\
(3) {\bf w/o $L_{rec}$}. Reconstruction loss is removed, essentially making the model a classification based residual network.\\ 
(4) {\bf w/o $L_{cls}$}. Taking inspiration from \cite{ImageGPT}, we first train the hourglass solely with the reconstruction task (\emph{i.e.} no $L_{cls}$) and then use the encoder module as a feature extractor to train the prediction block. For all the model variations, we keep everything else the same for consistency.

As presented in Table \ref{Tab:Ablation}, ablating the different components generates slightly different results. Due to the pixel normalization between $[-1, 1]$ before training, using MSE based reconstruction slightly dampens the error due to squaring. The skip connections help in propagating stronger signals across the network \cite{UNet}, hence removing them negatively affects model performance. Removing $L_{rec}$ altogether deteriorates model performance as contextual information gets overlooked. Surprisingly, unsupervised pre-training performs quite well, suggesting reconstruction can teach the model features useful for classification. This reconstruction element helps the full model achieve the best overall performance.

\vspace{-0.2cm}
\section{Conclusion}
In this work, we proposed a model that takes as input a patch of the driver's face along with a crop of the eye-region and provides a classification into 6 coarse ROIs in the vehicle. We demonstrated that an hourglass network consisting of encoder-decoder modules, trained with a secondary reconstruction loss, allows the model to learn strong feature representations and perform better in the primary glance classification task. In order to make the system more robust to subject-specific variations in appearance and driving behavior, we proposed a multi-stream model that takes a representation of a driver's baseline glance behavior as an auxiliary input for learning residuals. Results indicate such personalized training to improve model performance for multiple glance ROIs over rigid models. 

Finally, we designed a multi-domain training regime to jointly train our hourglass model on data collected from multiple camera views. Leveraging the hourglass' auxiliary reconstruction objective, this approach can learn domain invariant representations from very little labeled data in a weakly supervised manner, and consequently reduce labeling cost. As a future work, we plan to use our hourglass model as a proxy for annotating unlabeled data from new domains and actively learn from high confidence samples. 

\noindent {\bf Acknowledgements}: The AVT and MIT 2013 dataset used in this study were drawn from work supported by the Advanced Vehicle Technologies (AVT) Consortium at MIT (\url{http://agelab.mit.edu/avt}) and the Insurance Institute for Highway Safety (IIHS) respectively.

\section{Detailed Model Architecture}
Here we describe in detail the architecture of the encoder $E$ and decoder $D$ modules, and the prediction head $P$ of our two-channel hourglass model. As discussed in Section 4.1 of the main text, $E$ takes as input a 96$\times$96$\times$2 input and passes it through a dilated convolution layer \cite{atrous} before followed by 5 residual blocks \cite{ResNet} with stride = 2 for downsampling. This downsampled output is fed to a densely connected layer with 512 neurons and linear activation to generate the encoded feature representation of the input. $D$ is designed like a mirror image of $E$ and takes this dense 512-D input and feeds it through 5 upsampling pixel shuffling layers \cite{pixshuff}. We also add skip connections \cite{UNet} between layers in $E$ and $D$ with the same feature map resolution for stronger signal propagation. The final upsampled output is passed through a convolution layer with \emph{tanh} activation to reconstruct the 96$\times$96$\times$2 input \cite{DCGAN,salimans}. $P$ is composed of two dense layers with a dropout \cite{Dropout} layer in between for regularization. We apply softmax activation for the second dense layer to get the final glance prediction. Unless stated otherwise, all layers use a leaky \emph{ReLU} activation. 

The detailed layers of $E$, $D$ and $P$ are listed in Tables \ref{Tab:Encoder}, \ref{Tab:Decoder}, and \ref{Tab:Pred_Block} respectively. The convolution layers, dense layers, residual blocks and pixel shuffling blocks are represented as `conv', `fc', `RB', and `PS' respectively in the tables.

\begin{table}
\begin{center}
\captionsetup{justification=centering}
\caption{Encoder $E$ architecture (input size is 96$\times$96$\times$2)}
\begin{small}
\begin{tabular}{  | c | c| c| }
\hline
\begin{tabular}[x]{@{}c@{}}{\bf Layer}\end{tabular} & \begin{tabular}[x]{@{}c@{}}{\bf Filter/Stride/Dilation}\end{tabular} & \begin{tabular}[x]{@{}c@{}}{\bf \# of filters}\end{tabular}\\
\hline
\hline
  conv1 & 3$\times$3/1/2 & 128 \\
  conv2 & 3$\times$3/2/1 & 64\\
  RB1 & 3$\times$3/1/1 & 64\\
  conv3 & 3$\times$3/2/1 & 128\\
  RB2 & 3$\times$3/1/1 & 128\\
  conv4 & 3$\times$3/2/1 & 256\\
  RB3 & 3$\times$3/1/1 & 256\\
  conv5 & 3$\times$3/2/1 & 512\\
  RB4 & 3$\times$3/1/1 & 512\\
  conv6 & 3$\times$3/2/1 & 1,024 \\
  RB5 & 3$\times$3/1/1 & 1,024\\
  fc1 & 512 & - \\
  \hline
\end{tabular}
\label{Tab:Encoder}
\end{small}
\end{center}
\end{table}

\begin{table}
\begin{center}
\captionsetup{justification=centering}
\caption{Decoder $D$ architecture (input size is (512,)}
\begin{small}
\begin{tabular}{  | c | c| c| }
\hline
\begin{tabular}[x]{@{}c@{}}{\bf Layer}\end{tabular} & \begin{tabular}[x]{@{}c@{}}{\bf Filter/Stride/Dilation}\end{tabular} & \begin{tabular}[x]{@{}c@{}}{\bf \# of filters}\end{tabular}\\
\hline
\hline
  fc2 & 3*3*1024 & - \\ 
  conv7 & 3$\times$3/1/1 & 4*512\\
  PS1 & - & - \\
  conv8 & 3$\times$3/1/1 & 4*256\\
  PS2 & - & - \\
  conv9 & 3$\times$3/1/1 & 4*128\\
  PS3 & - & - \\
  conv10 & 3$\times$3/1/1 & 4*64\\
  PS4 & - & - \\
  conv11 & 3$\times$3/1/1 & 4*64\\
  PS5 & - & - \\
  conv12 & 5$\times$5/1/1 & 2\\
  \hline
\end{tabular}
\label{Tab:Decoder}
\end{small}
\end{center}
\end{table}

\begin{table}
\begin{center}
\captionsetup{justification=centering}
\caption{Prediction head $P$ architecture (input size is (512,)}
\begin{small}
\begin{tabular}{  | c | c| c| }
\hline
\begin{tabular}[x]{@{}c@{}}{\bf Layer}\end{tabular} & \begin{tabular}[x]{@{}c@{}}{\bf Filter/Stride/Dilation}\end{tabular} & \begin{tabular}[x]{@{}c@{}}{\bf \# of filters}\end{tabular}\\
\hline
\hline
  fc3 & 256 & - \\ 
  fc4 & 6 & - \\
  \hline
\end{tabular}
\label{Tab:Pred_Block}
\end{small}
\end{center}
\end{table}

\section{Classwise Data Distribution}
\begin{figure*}
\centering
   \includegraphics[width=1.0\linewidth]{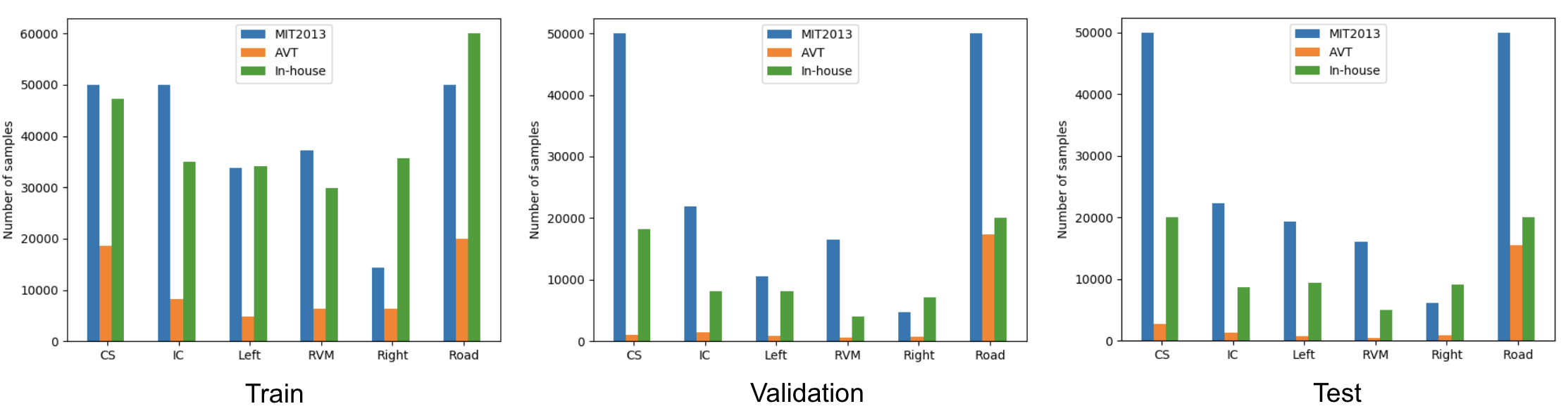}
   \caption{Class wise distribution of samples in the Train, Validation and Test splits in the MIT2013 \cite{mehler2016multi}, AVT \cite{fridman2019advanced} and our In-house datasets. The `Centerstack', `Instrument Cluster' and `Rearview Mirror' are abbreviated as `CS', `IC' and `RVM'.}
\label{fig:Data_Dist}
\end{figure*}

Here we present the class wise distribution of samples for the MIT2013 \cite{mehler2016multi}, AVT \cite{fridman2019advanced} and our In-house collected datasets in Figure \ref{fig:Data_Dist}. As can be seen, the pre-dominant class (ROI) is the driver actually looking on the road (`Road'), especially for the MIT2013 and In-house datasets. This imbalance can cause the trained model's representations to be skewed towards the largely populated ROIs and perform poorly for the sparse classes. However, as presented in the results from the main text our model does not exhibit such bias and performs competitively for all the ROI classes.

\begin{table*}
\begin{center}
\captionsetup{justification=centering}
\caption{Class-wise performance (ROC-AUC) of the different glance classification models on the MIT2013 dataset. This table is complementary to Table I in the main text.}
\begin{footnotesize}
\begin{tabular}{  | c| c| c| c| c| c| c| }
\hline
\begin{tabular}[x]{@{}c@{}}{\bf Model}\end{tabular} & \begin{tabular}[x]{@{}c@{}}{\bf Centerstack}\end{tabular} & \begin{tabular}[x]{@{}c@{}}{\bf Instrument}\\{\bf Cluster}\end{tabular} & \begin{tabular}[x]{@{}c@{}}{\bf Left}\end{tabular} & \begin{tabular}[x]{@{}c@{}}{\bf Rearview}\\{\bf Mirror}\end{tabular} & 
\begin{tabular}[x]{@{}c@{}}{\bf Right}\end{tabular} & 
\begin{tabular}[x]{@{}c@{}}{\bf Road}\end{tabular} \\
\hline
\hline
  \begin{tabular}[x]{@{}c@{}}Landmarks + MLP \cite{fridman2016owl}\end{tabular} & 0.903 $\pm$ 0.003 & 0.897 $\pm$ 0.011 & 0.956 $\pm$ 0.001 & 0.918 $\pm$ 0.001 & 0.883 $\pm$ 0.003 & 0.836 $\pm$ 0.004 \\
  \hline
  \begin{tabular}[x]{@{}c@{}}Baseline CNN \cite{AlexNet}\end{tabular} & 0.976 $\pm$ 0.001 & 0.939 $\pm$ 0.004 & 0.968 $\pm$ 0.003 & 0.975 $\pm$ 0.001 & 0.913 $\pm$ 0.002 & 0.946 $\pm$ 0.003 \\
  \hline
    \begin{tabular}[x]{@{}c@{}}Dari \emph{et al.} \cite{dari_gaze}\end{tabular} & 0.777 $\pm$ 0.001 & 0.742 $\pm$ 0.002 & 0.958 $\pm$ 0.001 & 0.763 $\pm$ 0.012 & 0.750 $\pm$ 0.001 & 0.743 $\pm$ 0.002 \\
  \hline
  \begin{tabular}[x]{@{}c@{}}Vora \emph{et al.} \cite{vora2018driver}\end{tabular} & 0.977 $\pm$ 0.002 & 0.949 $\pm$ 0.003 & 0.972 $\pm$ 0.003 & 0.979 $\pm$ 0.001 & 0.924 $\pm$ 0.007 & 0.959 $\pm$ 0.001 \\
  \hline
  \begin{tabular}[x]{@{}c@{}}Vora \emph{et al.} \cite{vora2018driver} + Decoder \end{tabular} & 0.982 $\pm$ 0.001 & 0.955 $\pm$ 0.003 & 0.976 $\pm$ 0.001 & {\bf 0.984 $\pm$ 0.001} & 0.935 $\pm$ 0.002 & 0.955 $\pm$ 0.003 \\
  \hline
  \begin{tabular}[x]{@{}c@{}}One-Channel Hourglass\end{tabular} & 0.982 $\pm$ 0.001 & 0.950 $\pm$ 0.003 & 0.975 $\pm$ 0.003 & 0.981 $\pm$ 0.001 & 0.927 $\pm$ 0.003 & 0.952 $\pm$ 0.003 \\
  \hline
  \begin{tabular}[x]{@{}c@{}}Two-Channel Hourglass (ours)\end{tabular} & {\bf 0.984 $\pm$ 0.001} & {\bf 0.959 $\pm$ 0.002} & {\bf 0.983 $\pm$ 0.002} & 0.982 $\pm$ 0.001 & 0.918 $\pm$ 0.007 & {\bf 0.969 $\pm$ 0.001} \\
\hline
  \begin{tabular}[x]{@{}c@{}}Personalized Hourglass (ours)\end{tabular} & {\bf 0.984 $\pm$ 0.001} & {\bf 0.959 $\pm$ 0.003} & 0.980 $\pm$ 0.003 & 0.983 $\pm$ 0.002 & {\bf 0.937 $\pm$ 0.003} & 0.961 $\pm$ 0.003 \\
  \hline
\end{tabular}
\label{Tab:MIT_Res_Full}
\end{footnotesize}
\end{center}
\vspace{-0.3cm}
\end{table*}

\begin{table*}
\begin{center}
\captionsetup{justification=centering}
\caption{Classwise breakdown of multi-domain ($d1$ = MIT2013, $d2$ = AVT) performance (ROC-AUC) of our hourglass network, trained using different regimes, on the MIT2013 dataset. This table corresponds to Table II in the main text.}
\begin{small}
\begin{tabular}{  | c| c| c| c| c| c| c| }
\hline
\begin{tabular}[x]{@{}c@{}}{\bf Model}\end{tabular} & \begin{tabular}[x]{@{}c@{}}{\bf Centerstack}\end{tabular} & \begin{tabular}[x]{@{}c@{}}{\bf Instrument}\\{\bf Cluster}\end{tabular} & \begin{tabular}[x]{@{}c@{}}{\bf Left}\end{tabular} & \begin{tabular}[x]{@{}c@{}}{\bf Rearview}\\{\bf Mirror}\end{tabular} & 
\begin{tabular}[x]{@{}c@{}}{\bf Right}\end{tabular} & 
\begin{tabular}[x]{@{}c@{}}{\bf Road}\end{tabular} \\
\hline
\hline
  \begin{tabular}[x]{@{}c@{}}Mixed Training\end{tabular} & 0.980 $\pm$ 0.001 &{\bf  0.956 $\pm$ 0.003} & {\bf 0.984 $\pm$ 0.001} & 0.979 $\pm$ 0.002 & 0.916 $\pm$ 0.003 & 0.965 $\pm$ 0.002 \\
  \hline
    \begin{tabular}[x]{@{}c@{}}Fine-tuning \cite{ImageGPT}\end{tabular} & 0.893 $\pm$ 0.001 & 0.768 $\pm$ 0.002 & 0.907 $\pm$ 0.003 & 0.909 $\pm$ 0.001 & 0.871 $\pm$ 0.001 & 0.903 $\pm$ 0.002 \\
  \hline
    \begin{tabular}[x]{@{}c@{}}Gradient Reversal \cite{GradRev}\end{tabular} & 0.976 $\pm$ 0.003 & 0.955 $\pm$ 0.001 & 0.979 $\pm$ 0.003 & 0.976 $\pm$ 0.003 & 0.914 $\pm$ 0.002 & {\bf 0.969 $\pm$ 0.002} \\
  \hline
  \begin{tabular}[x]{@{}c@{}}Tri-training \cite{tri_train_ruder}\end{tabular} & 0.976 $\pm$ 0.002 & 0.944 $\pm$ 0.004 & 0.971 $\pm$ 0.002 & 0.974 $\pm$ 0.002 & 0.914 $\pm$ 0.003 & 0.960 $\pm$ 0.001 \\
  \hline 
  \begin{tabular}[x]{@{}c@{}}Distillation \cite{distillation}\end{tabular} & 0.981 $\pm$ 0.001 & 0.954 $\pm$ 0.002 & 0.969 $\pm$ 0.002 & 0.978 $\pm$ 0.001 & {\bf 0.921 $\pm$ 0.001} & 0.962 $\pm$ 0.002 \\
  \hline
    \begin{tabular}[x]{@{}c@{}}Ours\end{tabular} & {\bf 0.983 $\pm$ 0.002} & 0.951 $\pm$ 0.003 & 0.980 $\pm$ 0.001 & {\bf 0.985 $\pm$ 0.001} & 0.918 $\pm$ 0.003 & 0.967 $\pm$ 0.001 \\
  \hline
\end{tabular}
\label{Tab:MIT_Dom_1_Full}
\end{small}
\end{center}
\vspace{-0.5cm}
\end{table*}

\begin{table*}
\begin{center}
\captionsetup{justification=centering}
\caption{Classwise breakdown of multi-domain ($d1$ = MIT2013, $d2$ = AVT) performance (ROC-AUC) of our hourglass network, trained using different regimes, on the AVT dataset. This table corresponds to Table II in the main text.}
\begin{small}
\begin{tabular}{  | c| c| c| c| c| c| c| }
\hline
\begin{tabular}[x]{@{}c@{}}{\bf Model}\end{tabular} & \begin{tabular}[x]{@{}c@{}}{\bf Centerstack}\end{tabular} & \begin{tabular}[x]{@{}c@{}}{\bf Instrument}\\{\bf Cluster}\end{tabular} & \begin{tabular}[x]{@{}c@{}}{\bf Left}\end{tabular} & \begin{tabular}[x]{@{}c@{}}{\bf Rearview}\\{\bf Mirror}\end{tabular} & 
\begin{tabular}[x]{@{}c@{}}{\bf Right}\end{tabular} & 
\begin{tabular}[x]{@{}c@{}}{\bf Road}\end{tabular} \\
\hline
\hline
  \begin{tabular}[x]{@{}c@{}}Mixed Training\end{tabular} & 0.963 $\pm$ 0.002 & 0.728 $\pm$ 0.017 & {\bf 0.958 $\pm$ 0.005} & 0.943 $\pm$ 0.006 & 0.972 $\pm$ 0.005 & 0.945 $\pm$ 0.004 \\
  \hline    
  \begin{tabular}[x]{@{}c@{}}Fine-tuning \cite{ImageGPT}\end{tabular} & {\bf 0.973 $\pm$ 0.002} & 0.741 $\pm$ 0.002 & 0.952 $\pm$ 0.003 & 0.941 $\pm$ 0.002 & 0.969 $\pm$ 0.003 & {\bf 0.950 $\pm$ 0.001} \\
  \hline
    \begin{tabular}[x]{@{}c@{}}Gradient Reversal \cite{GradRev}\end{tabular} & 0.972 $\pm$ 0.001 & 0.719 $\pm$ 0.002 & 0.945 $\pm$ 0.002 & 0.928 $\pm$ 0.005 & 0.970 $\pm$ 0.001 & 0.941 $\pm$ 0.002 \\
  \hline
  \begin{tabular}[x]{@{}c@{}}Tri-training \cite{tri_train_ruder}\end{tabular} & 0.960 $\pm$ 0.008 & {\bf 0.834 $\pm$ 0.016} & 0.941 $\pm$ 0.006 & {\bf 0.947 $\pm$ 0.001} & 0.969 $\pm$ 0.003 & 0.919 $\pm$ 0.009 \\
  \hline
  \begin{tabular}[x]{@{}c@{}}Distillation \cite{distillation}\end{tabular} & 0.943 $\pm$ 0.002 & 0.825 $\pm$ 0.002 & 0.954 $\pm$ 0.001 & 0.865 $\pm$ 0.003 & 0.931 $\pm$ 0.001 & 0.854 $\pm$ 0.002 \\
  \hline
    \begin{tabular}[x]{@{}c@{}}Ours\end{tabular} & 0.964 $\pm$ 0.001 & 0.736 $\pm$ 0.002 & 0.948 $\pm$ 0.001 & {\bf 0.947 $\pm$ 0.002} & {\bf 0.977 $\pm$ 0.002} & 0.940 $\pm$ 0.001 \\
  \hline
\end{tabular}
\label{Tab:AVT_Dom_1_Full}
\end{small}
\end{center}
\vspace{-0.5cm}
\end{table*}

\begin{table*}
\begin{center}
\captionsetup{justification=centering}
\caption{Classwise breakdown of multi-domain ($d1$ = MIT2013, $d2$ = In-house) performance (ROC-AUC) of our hourglass network, trained using different regimes, on the MIT2013 dataset. This table corresponds to Table III in the main text.}
\begin{small}
\begin{tabular}{  | c| c| c| c| c| c| c| }
\hline
\begin{tabular}[x]{@{}c@{}}{\bf Model}\end{tabular} & \begin{tabular}[x]{@{}c@{}}{\bf Centerstack}\end{tabular} & \begin{tabular}[x]{@{}c@{}}{\bf Instrument}\\{\bf Cluster}\end{tabular} & \begin{tabular}[x]{@{}c@{}}{\bf Left}\end{tabular} & \begin{tabular}[x]{@{}c@{}}{\bf Rearview}\\{\bf Mirror}\end{tabular} & 
\begin{tabular}[x]{@{}c@{}}{\bf Right}\end{tabular} & 
\begin{tabular}[x]{@{}c@{}}{\bf Road}\end{tabular} \\
\hline
\hline
  \begin{tabular}[x]{@{}c@{}}Mixed Training\end{tabular} & {\bf 0.978 $\pm$ 0.001} & 0.937 $\pm$ 0.007 & 0.978 $\pm$ 0.001 & {\bf 0.978 $\pm$ 0.001} & 0.912 $\pm$ 0.006 & 0.955 $\pm$ 0.003 \\
  \hline
    \begin{tabular}[x]{@{}c@{}}Fine-tuning \cite{ImageGPT}\end{tabular} & 0.808 $\pm$ 0.003 & 0.641 $\pm$ 0.003 & 0.927 $\pm$ 0.002 & 0.834 $\pm$ 0.002 & 0.833 $\pm$ 0.002 & 0.800 $\pm$ 0.001 \\
  \hline
    \begin{tabular}[x]{@{}c@{}}Gradient Reversal \cite{GradRev}\end{tabular} & 0.970 $\pm$ 0.002 & 0.929 $\pm$ 0.001 & 0.959 $\pm$ 0.001 & 0.970 $\pm$ 0.001 & 0.907 $\pm$ 0.005 & 0.903 $\pm$ 0.006 \\
  \hline
  \begin{tabular}[x]{@{}c@{}}Tri-training \cite{tri_train_ruder}\end{tabular} & 0.971 $\pm$ 0.002 & 0.932 $\pm$ 0.004 & 0.972 $\pm$ 0.002 & 0.972 $\pm$ 0.001 & 0.915 $\pm$ 0.003 & 0.959 $\pm$ 0.002 \\
  \hline
  \begin{tabular}[x]{@{}c@{}}Distillation \cite{distillation}\end{tabular} & 0.964 $\pm$ 0.001 & 0.940 $\pm$ 0.002 & 0.974 $\pm$ 0.001 & 0.973 $\pm$ 0.001 & 0.869 $\pm$ 0.002 & 0.951 $\pm$ 0.002 \\
  \hline
    \begin{tabular}[x]{@{}c@{}}Ours\end{tabular} & 0.974 $\pm$ 0.002 & {\bf 0.957 $\pm$ 0.002} & {\bf 0.979 $\pm$ 0.002} & 0.976 $\pm$ 0.002 & {\bf 0.920 $\pm$ 0.001} & {\bf 0.965 $\pm$ 0.002} \\
  \hline
\end{tabular}
\label{Tab:MIT_Dom_2_Full}
\end{small}
\end{center}
\vspace{-0.5cm}
\end{table*}

\begin{table*}
\begin{center}
\captionsetup{justification=centering}
\caption{Classwise breakdown of multi-domain ($d1$ = MIT2013, $d2$ = In-house) performance (ROC-AUC) of our hourglass network, trained using different regimes, on the In-house dataset. This table corresponds to Table III in the main text.}
\begin{small}
\begin{tabular}{  | c| c| c| c| c| c| c| }
\hline
\begin{tabular}[x]{@{}c@{}}{\bf Model}\end{tabular} & \begin{tabular}[x]{@{}c@{}}{\bf Centerstack}\end{tabular} & \begin{tabular}[x]{@{}c@{}}{\bf Instrument}\\{\bf Cluster}\end{tabular} & \begin{tabular}[x]{@{}c@{}}{\bf Left}\end{tabular} & \begin{tabular}[x]{@{}c@{}}{\bf Rearview}\\{\bf Mirror}\end{tabular} & 
\begin{tabular}[x]{@{}c@{}}{\bf Right}\end{tabular} & 
\begin{tabular}[x]{@{}c@{}}{\bf Road}\end{tabular} \\
\hline
\hline
  \begin{tabular}[x]{@{}c@{}}Mixed Training\end{tabular} & 0.881 $\pm$ 0.002 & 0.775 $\pm$ 0.005 & 0.929 $\pm$ 0.003 & {\bf 0.928 $\pm$ 0.004} & 0.949 $\pm$ 0.002 & 0.834 $\pm$ 0.006 \\
  \hline
    \begin{tabular}[x]{@{}c@{}}Fine-tuning \cite{ImageGPT}\end{tabular} & {\bf 0.891 $\pm$ 0.002} & 0.721 $\pm$ 0.002 & 0.892 $\pm$ 0.002 & 0.914 $\pm$ 0.001 & 0.930 $\pm$ 0.003 & 0.800 $\pm$ 0.005 \\
  \hline
    \begin{tabular}[x]{@{}c@{}}Gradient Reversal \cite{GradRev}\end{tabular} & 0.827 $\pm$ 0.004 & {\bf 0.828 $\pm$ 0.004} & 0.868 $\pm$ 0.002 & 0.872 $\pm$ 0.004 & 0.925 $\pm$ 0.005 & 0.778 $\pm$ 0.003 \\
  \hline
  \begin{tabular}[x]{@{}c@{}}Tri-training \cite{tri_train_ruder}\end{tabular} & 0.870 $\pm$ 0.007 & 0.676 $\pm$ 0.008 & 0.884 $\pm$ 0.012 & 0.886 $\pm$ 0.015 & 0.934 $\pm$ 0.006 & 0.789 $\pm$ 0.008 \\
  \hline
  \begin{tabular}[x]{@{}c@{}}Distillation \cite{distillation}\end{tabular} & 0.854 $\pm$ 0.002 & 0.661 $\pm$ 0.004 & {\bf 0.954 $\pm$ 0.003} & 0.838 $\pm$ 0.002 & 0.884 $\pm$ 0.002 & 0.789 $\pm$ 0.002 \\
  \hline
    \begin{tabular}[x]{@{}c@{}}Ours\end{tabular} & 0.886 $\pm$ 0.003 & 0.738 $\pm$ 0.008 & 0.901 $\pm$ 0.005 & 0.924 $\pm$ 0.003 & {\bf 0.962 $\pm$ 0.002} & {\bf 0.849 $\pm$ 0.005} \\
  \hline
\end{tabular}
\label{Tab:Aff_Dom_2_Full}
\end{small}
\end{center}
\vspace{-0.2cm}
\end{table*}

\begin{table*}
\begin{center}
\captionsetup{justification=centering}
\caption{Classwise performance (ROC-AUC) of our two-channel hourglass model with different components ablated on the MIT2013 dataset. This corresponds to Table V in the main text.}
\begin{small}
\begin{tabular}{  | c| c| c| c| c| c| c| }
\hline
\begin{tabular}[x]{@{}c@{}}{\bf Model}\end{tabular} & \begin{tabular}[x]{@{}c@{}}{\bf Centerstack}\end{tabular} & \begin{tabular}[x]{@{}c@{}}{\bf Instrument Cluster}\end{tabular} & \begin{tabular}[x]{@{}c@{}}{\bf Left}\end{tabular} & \begin{tabular}[x]{@{}c@{}}{\bf Rearview Mirror}\end{tabular} & 
\begin{tabular}[x]{@{}c@{}}{\bf Right}\end{tabular} & 
\begin{tabular}[x]{@{}c@{}}{\bf Road}\end{tabular} \\
\hline
\hline
  \begin{tabular}[x]{@{}c@{}}w/ MSE\end{tabular} & 0.981 $\pm$ 0.002 & 0.955 $\pm$ 0.003 & 0.982 $\pm$ 0.001 & 0.980 $\pm$ 0.002 & 0.917 $\pm$ 0.006 & 0.964 $\pm$ 0.002 \\
  \hline
  \begin{tabular}[x]{@{}c@{}}wo/ skip connections\end{tabular} & 0.979 $\pm$ 0.001 & 0.955 $\pm$ 0.007 & 0.982 $\pm$ 0.002 & 0.981 $\pm$ 0.002 & 0.910 $\pm$ 0.008 & 0.960 $\pm$ 0.003 \\
  \hline
  \begin{tabular}[x]{@{}c@{}}wo/ $L_{rec}$\end{tabular} & 0.978 $\pm$ 0.002 & 0.943 $\pm$ 0.002 & 0.977 $\pm$ 0.001 & 0.979 $\pm$ 0.001 & 0.933 $\pm$ 0.004 & 0.945 $\pm$ 0.003 \\
  \hline
  \begin{tabular}[x]{@{}c@{}}wo/ $L_{cls}$\cite{ImageGPT}\end{tabular} & 0.982 $\pm$ 0.002 & 0.954 $\pm$ 0.002 & 0.979 $\pm$ 0.001 & 0.981 $\pm$ 0.002 & 0.927 $\pm$ 0.003 & 0.951 $\pm$ 0.002 \\
  \hline
  \begin{tabular}[x]{@{}c@{}}Full Model\end{tabular} & 0.984 $\pm$ 0.001 & 0.959 $\pm$ 0.002 & 0.983 $\pm$ 0.002 & 0.982 $\pm$ 0.001 & 0.918 $\pm$ 0.007 & 0.969 $\pm$ 0.001 \\
  \hline
\end{tabular}
\label{Tab:Ablation_Full}
\end{small}
\end{center}
\vspace{-0.2cm}
\end{table*}

\section{Classwise Breakdown of Model Performance \& Statistical Significance}
To compute statistical significance between the performance of various methods, we train each model 5 times using random seeds for initialization, and applied a one-tailed paired t-test on their macro-average ROC-AUCs. Comparing the classification results of our proposed two-channel hourglass model with others on the MIT2013 dataset (Table I of main paper), we found a statistically significant difference when compared with:
\begin{itemize}
  \item the Landmarks + MLP model (p = 0.0000001 $<$ 0.05),
  \item the Dense Eye Landmarks + Headpose model (p = 0.000000005 $<$ 0.05),
  \item the Baseline CNN model (p = 0.0001 $<$ 0.05),
  \item the Upperface SqueezeNet model (p = 0.008 $<$ 0.05), and
  \item the One-channel Hourglass model (p = 0.002 $<$ 0.05),
\end{itemize}

Comparing the classification results of our proposed personalized hourglass model with others on the MIT2013 dataset (Table I of main paper), we found a statistically significant difference when compared with:
\begin{itemize}
  \item the Landmarks + MLP model (p = 0.00000003 $<$ 0.05),
  \item the Dense Eye Landmarks + Headpose model (p = 0.00000002 $<$ 0.05),
  \item the Baseline CNN model (p = 0.00009 $<$ 0.05),
  \item the Upperface SqueezeNet model (p = 0.0004 $<$ 0.05),
  \item the One-channel Hourglass model (p = 0.001 $<$ 0.05), and
  \item the Upperface SqueezeNet model w/ Decoder (p = 0.003 $<$ 0.05),
\end{itemize}

Comparing the results of our proposed multi-domain model on the target domain (AVT) (Table II of main paper), we found a statistically significant difference when compared with:
\begin{itemize}
\item Gradient Reversal (p = 0.0007 $<$ 0.05), and
\item Distillation (p = 0.000003 $<$ 0.05)
\end{itemize}

Comparing the results of our proposed multi-domain model on the target domain (In-house) (Table III of main paper), we found a statistically significant difference when compared with:
\begin{itemize}
\item Fine-tuning (p = 0.0003 $<$ 0.05), 
\item Gradient Reversal (p = 0.000008 $<$ 0.05), 
\item Tri-training (p = 0.000004 $<$ 0.05), and
\item Distillation (p = 0.00001 $<$ 0.05)
\end{itemize}

Quantifying the effects of adding the various components of our full model (Table V of main paper), we found a statistically significant different when: 
\begin{itemize}
\item comparing the baseline model without the decoder with an hourglass model with a decoder without skip connections (p = 0.01 $<$ 0.05), and
\item comparing the hourglass model with the decoder without the skip connections with our final model (p = 0.01 $<$ 0.05).
\end{itemize}

Similarly, comparing our final model with a two-stage model (pre-trained unsupervised encoder fine-tuned with classification loss) was found to be statistically significant (p = 0.0004 $<$ 0.05). Comparing our final model with a model trained with MSE (instead of MAE) also yielded statistically significant results (p = 0.01 $<$ 0.05).

\begin{figure*}[t]
\centering
   \includegraphics[width=1.0\linewidth]{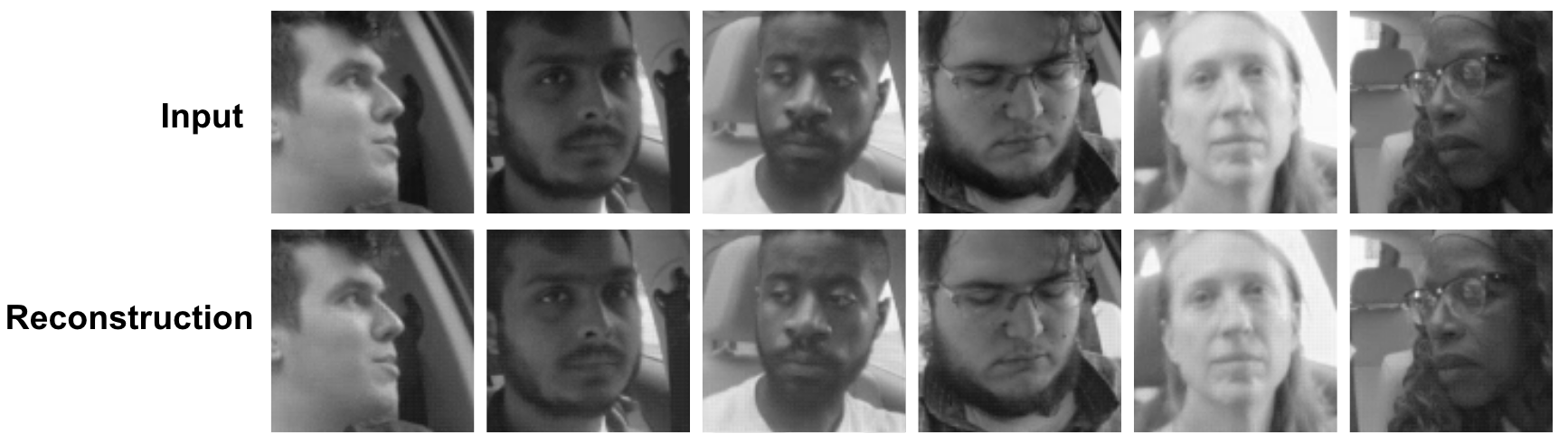}
   \caption{Randomly sampled cropped face images (input) and their reconstructions produced by our hourglass model. Subtle differences between the two sets can be observed by zooming in. All images are 96$\times$96$\times$1 in resolution.}
\label{fig:Recon_Ex}
\end{figure*}

\section{Reconstruction Example}
We present a random set of face samples from our In-house dataset and their reconstructions generated by our hourglass model in Figure \ref{fig:Recon_Ex}. Except for some noise and grid-like artifact in some cases, we find there to be little difference between the input and reconstructed images.

\begin{figure*}[t]
\centering
  \includegraphics[width=1.0\linewidth]{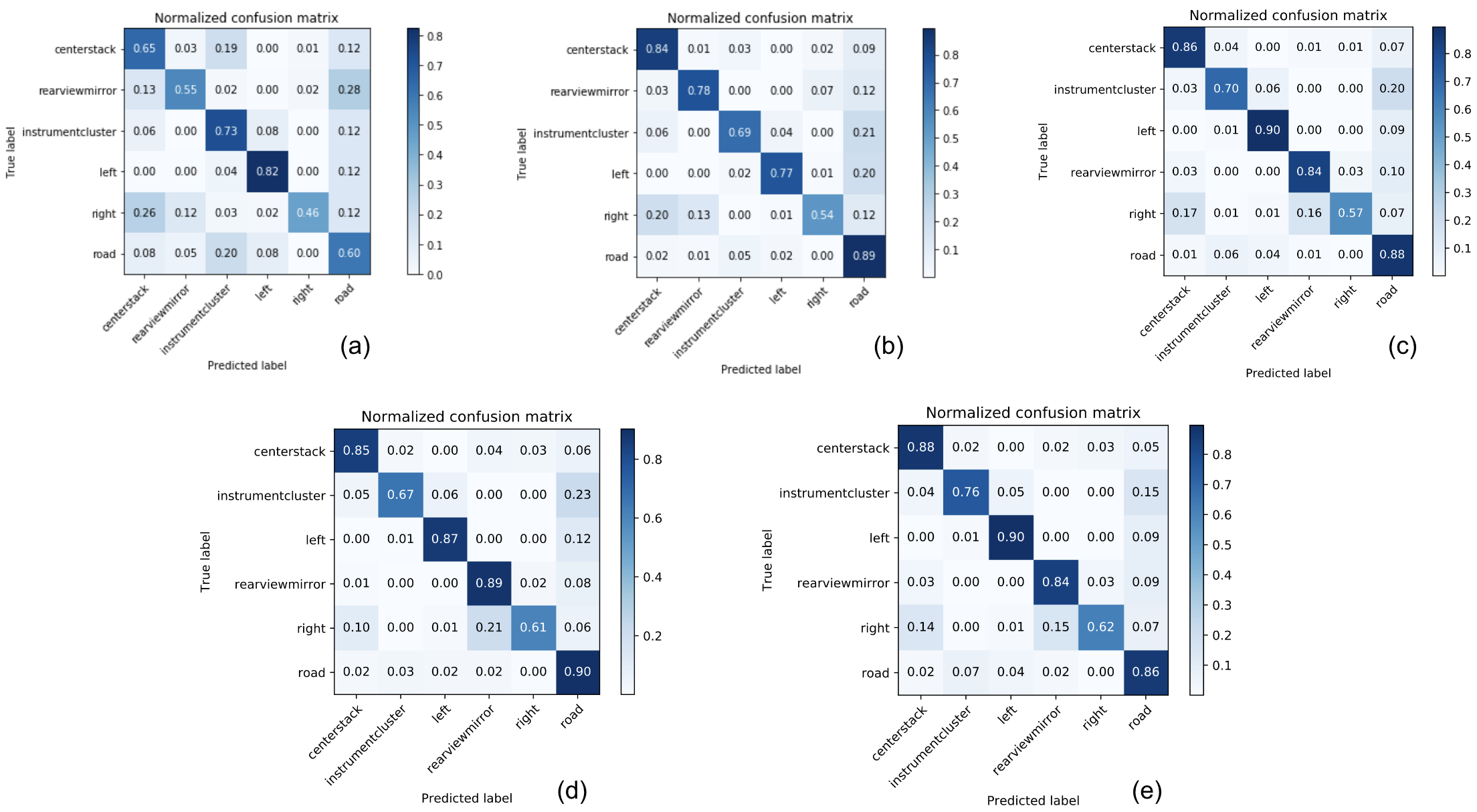}
  \caption{Normalized confusion matrices across the different classes on the MIT2013 \cite{mehler2016multi} test samples using - (a) Landmarks + MLP model, (b) Baseline CNN \cite{AlexNet}, (c) One-Channel Hourglass, (d) Two-Channel Hourglass, and (e) Personalized Hourglass model.}
\label{fig:Con_Mats}
\end{figure*}

\section{Confusion Matrices}
In the main text, we utilize ROC-AUC as the metric to report performance of the different models. Since the ROC-AUC metric is threshold agnostic, it can be used to gauge model performance while sweeping through different thresholds. However, we also present the performance of each of our candidate models on the MIT2013 \cite{mehler2016multi} test set in Figure \ref{fig:Con_Mats} using confusion matrices normalized by total samples for each class.

\section{Feature Visualization}
\begin{figure*}[t]
\centering
  \includegraphics[width=1.0\linewidth]{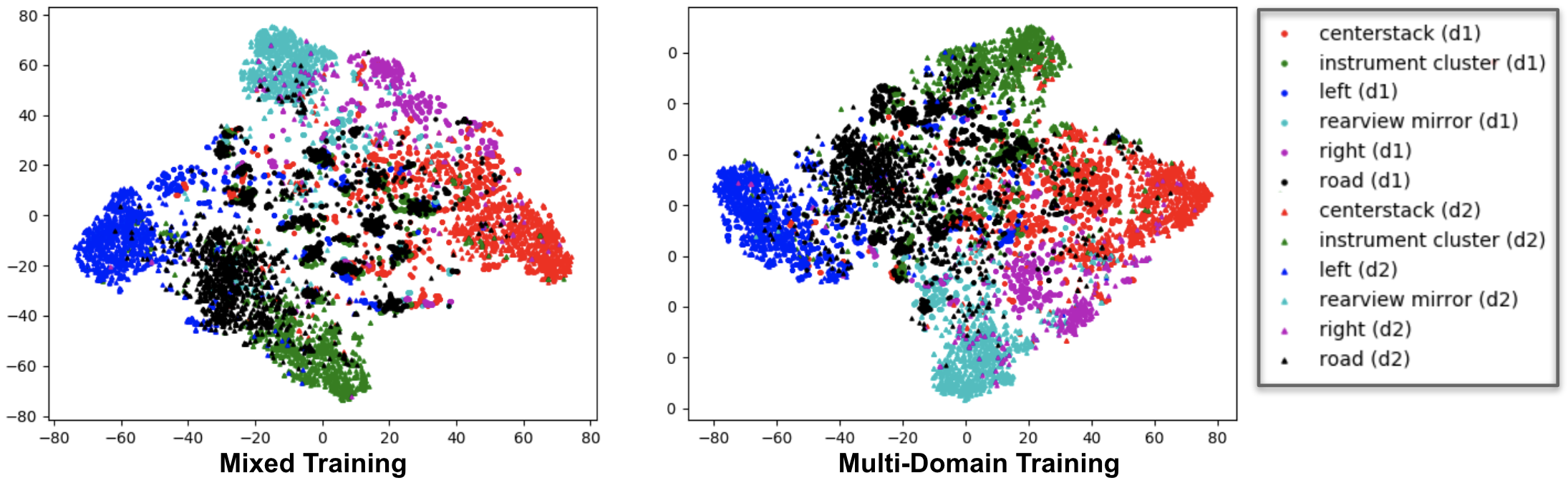}
  \caption{tSNE \cite{tsne} visualization of the encoded features: Our multi-domain training more compactly packs together the feature samples from $d1$ (MIT2013) and $d2$ (In-house dataset) than mixed-training, especially for the critical `Road' class (in black).}
\label{fig:tSNE}
\vspace{-0.5cm}
\end{figure*}

We visualize the encoded class-wise features using tSNE \cite{tsne}, as depicted in Figure \ref{fig:tSNE}. Our multi-domain training packs together the feature samples from $d1$ and $d2$ more compactly than mixed-training, especially for the critical `Road' class. Thus, this technique can be used to gauge the amount of labeling required when adapting models to new domains and consequently reduce annotation cost.

\vspace{-0.3cm}

{\small
\bibliographystyle{ieee_fullname}
\bibliography{egbib}

\begin{thebibliography}{10}\itemsep=-1pt

\bibitem{tobii}
Tobii.
\newblock \url{https://www.tobii.com/}.

\bibitem{TF}
M. Abadi et~al.
\newblock Tensorflow: A system for large-scale machine learning.
\newblock In {\em OSDI Symposia}, 2016.

\bibitem{SREFI3}
S. Banerjee et~al.
\newblock On hallucinating context and background pixels from a face mask using
  multi-scale gans.
\newblock In {\em WACV}, 2020.

\bibitem{SREFI2}
S. Banerjee, W. Scheirer, K. Bowyer, and P. Flynn.
\newblock Fast face image synthesis with minimal training.
\newblock In {\em WACV}, 2019.

\bibitem{PruningGuide}
D. Blalock, J.~J. Gonzalez~Ortiz, J. Frankle, and J. Guttag.
\newblock What is the state of neural network pruning?
\newblock In {\em MLSys}, 2020.

\bibitem{cazzato2020look}
D. Cazzato, M. Leo, C. Distante, and H. Voos.
\newblock When i look into your eyes: A survey on computer vision contributions
  for human gaze estimation and tracking.
\newblock {\em Sensors}, 20(13):3739, 2020.

\bibitem{chang2019salgaze}
Z. Chang et~al.
\newblock Salgaze: Personalizing gaze estimation using visual saliency.
\newblock In {\em ICCV Workshops}, 2019.

\bibitem{ImageGPT}
M. Chen, A. Radford, R. Child, J. Wu, H. Jun, D. Luan, and I. Sutskever.
\newblock Generative pretraining from pixels.
\newblock In {\em ICML}, 2020.

\bibitem{chhabra2017survey}
R. Chhabra, S. Verma, and C.R. Krishna.
\newblock A survey on driver behavior detection techniques for intelligent
  transportation systems.
\newblock In {\em Confluence}, 2017.

\bibitem{chollet2015keras}
F. Chollet et~al.
\newblock Keras.
\newblock \url{https://github.com/fchollet/keras}, 2015.

\bibitem{chu2013selective}
W.S. Chu, F. De~la Torre, and J. Cohn.
\newblock Selective transfer machine for personalized facial action unit
  detection.
\newblock In {\em CVPR}, 2013.

\bibitem{coughlin2011monitoring}
J. Coughlin, B. Reimer, and B. Mehler.
\newblock Monitoring, managing, and motivating driver safety and well-being.
\newblock {\em Pervasive Computing}, 10(3):14--21, 2011.

\bibitem{dari_gaze}
S. Dari, N. Kadrileev, and E. Hüllermeier.
\newblock A neural network-based driver gaze classification system with vehicle
  signals.
\newblock In {\em IJCNN}, 2020.

\bibitem{ArcFace}
J. Deng, J. Guo, N. Xue, and S. Zafeiriou.
\newblock Arcface: Additive angular margin loss for deep face recognition.
\newblock In {\em CVPR}, 2019.

\bibitem{FB_Reg}
X. Dong, S. Yu, X. Weng, S. Wei, Y. Yang, and Y. Sheikh.
\newblock Supervision-by-registration: An unsupervised approach to improve the
  precision of facial landmark detectors.
\newblock In {\em CVPR}, 2018.

\bibitem{fitch2013impact}
G. Fitch et~al.
\newblock The impact of hand-held and hands-free cell phone use on driving
  performance and safety-critical event risk.
\newblock Technical report, 2013.

\bibitem{fridman2019advanced}
L. Fridman et~al.
\newblock Mit advanced vehicle technology study: Large-scale naturalistic
  driving study of driver behavior and interaction with automation.
\newblock {\em IEEE Access}, 7:102021--102038, 2019.

\bibitem{fridman2016owl}
L. Fridman, J. Lee, B. Reimer, and T. Victor.
\newblock ‘owl’and ‘lizard’: Patterns of head pose and eye pose in
  driver gaze classification.
\newblock {\em IET Computer Vision}, 10(4):308--314, 2016.

\bibitem{fridman2018cognitive}
L. Fridman, B. Reimer, B. Mehler, and W. Freeman.
\newblock Cognitive load estimation in the wild.
\newblock In {\em CHI}, 2018.

\bibitem{GradRev}
Y. Ganin et~al.
\newblock Domain-adversarial training of neural networks.
\newblock {\em JMLR}, 17(59):1--35, 2016.

\bibitem{FineGrainedDom}
T. Gebru, J. Hoffman, and F-F. Li.
\newblock Fine-grained recognition in the wild: A multi-task domain adaptation
  approach.
\newblock In {\em ICCV}, 2017.

\bibitem{dgw}
S. Ghosh, A. Dhall, G. Sharma, S. Gupta, and N. Sebe.
\newblock Speak2label: Using domain knowledge for creating a large scale driver
  gaze zone estimation dataset.
\newblock {\em arXiv:2004.05973}.

\bibitem{hansen2009eye}
D.W. Hansen and Q. Ji.
\newblock In the eye of the beholder: A survey of models for eyes and gaze.
\newblock {\em T-PAMI}, 32(3):478--500, 2009.

\bibitem{HarisTTIC}
M. Haris, G. Shakhnarovich, and N. Ukita.
\newblock Task-driven super resolution: Object detection in low-resolution
  images.
\newblock {\em arXiv:1803.11316}.

\bibitem{ResNet}
K. He, X. Zhang, S. Ren, and J. Sun.
\newblock Deep residual learning for image recognition.
\newblock In {\em CVPR}, 2016.

\bibitem{distillation}
G. Hinton, O. Vinyals, and J. Dean.
\newblock Distilling the knowledge in a neural network.
\newblock {\em arXiv:1503.02531}.

\bibitem{TransferLearning}
M. Huh, P. Agarwal, and AA. Efros.
\newblock What makes imagenet good for transfer learning?
\newblock {\em arXiv:1608.08614}.

\bibitem{squeezenet}
FN. Iandola et~al.
\newblock Squeezenet: Alexnet-level accuracy with 50x fewer parameters and
  <0.5mb model size.
\newblock In {\em ICLR}, 2017.

\bibitem{Bulat3D}
A.~S. Jackson, A. Bulat, V. Argyriou, and G. Tzimiropoulos.
\newblock Large pose 3d face reconstruction from a single image via direct
  volumetric cnn regression.
\newblock {\em ICCV}, 2017.

\bibitem{jha2018probabilistic}
S. Jha and C. Busso.
\newblock Probabilistic estimation of the gaze region of the driver using dense
  classification.
\newblock In {\em ITSC}, pages 697--702, 2018.

\bibitem{joshi2017personalizing}
A. Joshi, S. Ghosh, M. Betke, S. Sclaroff, and H. Pfister.
\newblock Personalizing gesture recognition using hierarchical bayesian neural
  networks.
\newblock In {\em CVPR}, 2017.

\bibitem{joshi2020inthewild}
A. Joshi, S. Kyal, S. Banerjee, and T. Mishra.
\newblock In-the-wild drowsiness detection from facial expressions.
\newblock In {\em IV}, 2020.

\bibitem{kim2019nvgaze}
J. Kim, M. Stengel, A. Majercik, S. De~Mello, D. Dunn, S. Laine, M. McGuire,
  and D. Luebke.
\newblock Nvgaze: An anatomically-informed dataset for low-latency, near-eye
  gaze estimation.
\newblock In {\em CHI}, 2019.

\bibitem{Adam}
D. Kingma and J. Ba.
\newblock Adam: A method for stochastic optimization.
\newblock In {\em ICLR}, 2015.

\bibitem{klauer2006impact}
SG. Klauer et~al.
\newblock The impact of driver inattention on near-crash/crash risk: An
  analysis using the 100-car naturalistic driving study data.
\newblock 2006.

\bibitem{ImgNetTransfer}
S. Kornblith, J. Shlens, and QV. Le.
\newblock Do better imagenet models transfer better?
\newblock In {\em CVPR}, 2019.

\bibitem{krafka2016eye}
K. Krafka, A. Khosla, P. Kellnhofer, H. Kannan, S. Bhandarkar, W. Matusik, and
  A. Torralba.
\newblock Eye tracking for everyone.
\newblock In {\em CVPR}, 2016.

\bibitem{AlexNet}
A. Krizhevsky, I. Sutskever, and G.~E. Hinton.
\newblock Imagenet classification with deep convolutional neural networks.
\newblock In {\em NeurIPS}, 2012.

\bibitem{liang2012dangerous}
Y. Liang, J. Lee, and L. Yekhshatyan.
\newblock How dangerous is looking away from the road? algorithms predict crash
  risk from glance patterns in naturalistic driving.
\newblock {\em Human Factors}, 54(6):1104--1116, 2012.

\bibitem{linden2019learning}
E. Lind{\'e}n, J. Sjostrand, and A. Proutiere.
\newblock Learning to personalize in appearance-based gaze tracking.
\newblock In {\em ICCV Workshops}, 2019.

\bibitem{SphereFace}
W. Liu, Y. Wen, Z. Yu, M. Li, B. Raj, and S. Le.
\newblock Sphereface: Deep hypersphere embedding for face recognition.
\newblock In {\em CVPR}, 2017.

\bibitem{AdvAuto}
A. Makhzani, J. Shlens, N. Jaitly, and I. Goodfellow.
\newblock Adversarial autoencoders.
\newblock In {\em ICLR}, 2016.

\bibitem{PiggyBack}
A. Mallya, D. Davis, and S. Lazebnik.
\newblock Piggyback: Adapting a single network to multiple tasks by learning to
  mask weights.
\newblock In {\em ECCV}, 2018.

\bibitem{PackNet}
A. Mallya and S. Lazebnik.
\newblock Packnet: Adding multiple tasks to a single network by iterative
  pruning.
\newblock In {\em CVPR}, 2018.

\bibitem{MasiAug}
I. Masi, A.~T. Tran, J.~T. Leksut, T. Hassner, and G. Medioni.
\newblock Do we really need to collect millions of faces for effective face
  recognition?
\newblock In {\em ECCV}, 2016.

\bibitem{mehler2016multi}
B. Mehler et~al.
\newblock Multi-modal assessment of on-road demand of voice and manual phone
  calling and voice navigation entry across two embedded vehicle systems.
\newblock {\em Ergonomics}, 59(3):344--367, 2016.

\bibitem{morando2020driver}
A. Morando, P. Gershon, B. Mehler, and B. Reimer.
\newblock Driver-initiated tesla autopilot disengagements in naturalistic
  driving.
\newblock In {\em AutomotiveUI}, 2020.

\bibitem{NetTailor}
P. Morgado and N. Vasconcelos.
\newblock Nettailor: Tuning the architecture, not just the weights.
\newblock In {\em CVPR}, 2019.

\bibitem{DecoderTTIC}
M. Mostajabi, M. Maire, and G. Shakhnarovich.
\newblock Regularizing deep networks by modeling and predicting label
  structure.
\newblock In {\em CVPR}, 2018.

\bibitem{StackedHourGlass}
A. Newell, K. Yang, and J. Deng.
\newblock Stacked hourglass networks for human pose estimation.
\newblock In {\em ECCV}, 2016.

\bibitem{Peng2019DomainAL}
X. Peng, Z. Huang, X. Sun, and K. Saenko.
\newblock Domain agnostic learning with disentangled representations.
\newblock In {\em ICML}, 2019.

\bibitem{PJP_FG}
P.J. Phillips.
\newblock A cross benchmark assessment of a deep convolutional neural network
  for face recognition.
\newblock In {\em FG}, 2017.

\bibitem{DCGAN}
A. Radford, L. Metz, and S. Chintala.
\newblock Unsupervised representation learning with deep convolutional
  generative adversarial networks.
\newblock In {\em ICLR}, 2016.

\bibitem{rangesh2020driver}
A. Rangesh, B. Zhang, and M. Trivedi.
\newblock Driver gaze estimation in the real world: Overcoming the eyeglass
  challenge.
\newblock {\em arXiv:2002.02077}.

\bibitem{Context3}
A. Rice, P.J. Phillips, V. Natu, X. An, and A.J. O'Toole.
\newblock Unaware person recognition from the body when face identification
  fails.
\newblock {\em Psychological Science}, 24:2235--2243, 2013.

\bibitem{UNet}
O. Ronneberger, P. Fischer, and T. Brox.
\newblock U-net: Convolutional networks for biomedical image segmentation.
\newblock In {\em MICCAI}, 2015.

\bibitem{tri_train_ruder}
S. Ruder and B. Plank.
\newblock Strong baselines for neural semi-supervised learning under domain
  shift.
\newblock In {\em ACL}, 2018.

\bibitem{ILSVRC15}
O. Russakovsky et~al.
\newblock {ImageNet Large Scale Visual Recognition Challenge}.
\newblock {\em IJCV}, 115(3):211--252, 2015.

\bibitem{DA2010}
K. Saenko, B. Kulis, M. Fritz, , and T. Darrell.
\newblock Adapting visual category models to new domains.
\newblock In {\em ECCV}, 2010.

\bibitem{salimans}
T. Salimans, I. Goodfellow, W. Zaremba, V. Cheung, A. Radford, and X. Chen.
\newblock Improved techniques for training gans.
\newblock In {\em NeurIPS}, 2016.

\bibitem{seppelt2017glass}
B.D. Seppelt, S. Seaman, J. Lee, L.S. Angell, B. Mehler, and B. Reimer.
\newblock Glass half-full: On-road glance metrics differentiate crashes from
  near-crashes in the 100-car data.
\newblock {\em Accident Analysis \& Prevention}, 107:48--62, 2017.

\bibitem{SharmaTask}
V. Sharma, A. Diba, D. Neven, MS. Brown, L. Van~Gool, and R. Stiefelhagen.
\newblock Classification-driven dynamic image enhancement.
\newblock In {\em CVPR}, 2018.

\bibitem{pixshuff}
W. Shi et~al.
\newblock Real-time single image and video super-resolution using an efficient
  sub-pixel convolutional neural network.
\newblock In {\em CVPR}, 2016.

\bibitem{shrivastava2017learning}
A. Shrivastava, T. Pfister, O. Tuzel, J. Susskind, W. Wang, and R. Webb.
\newblock Learning from simulated and unsupervised images through adversarial
  training.
\newblock In {\em CVPR}, 2017.

\bibitem{vgg_sim}
K. Simonyan and A. Zisserman.
\newblock Very deep convolutional networks for large-scale image recognition.
\newblock In {\em ICLR}, 2015.

\bibitem{smith2005methodology}
D.L. Smith et~al.
\newblock Methodology for capturing driver eye glance behavior during
  in-vehicle secondary tasks.
\newblock {\em Transportation Research Record}, 1937(1):61--65, 2005.

\bibitem{Dropout}
N. Srivastava, G. Hinton, A. Krizhevsky, I. Sutskever, and R. Salakhutdinov.
\newblock Dropout: A simple way to prevent neural networks from overfitting.
\newblock {\em JMLR}, 15(56):1929--1958, 2014.

\bibitem{GradRevDrive}
M. Tonutti, E. Ruffaldi, A. Cattaneo, and CA. Avizzano.
\newblock Robust and subject-independent driving manoeuvre anticipation through
  domain-adversarial recurrent neural networks.
\newblock {\em Robotics and Autonomous Systems}, 115:162--173, 2019.

\bibitem{SimDeepTrans}
E. Tzeng, J. Hoffman, T. Darrell, and K. Saenko.
\newblock Simultaneous deep transfer across domains and tasks.
\newblock In {\em ICCV}, 2015.

\bibitem{AdvDiscDA}
E. Tzeng, J. Hoffman, K. Saenko, and T. Darrell.
\newblock Adversarial discriminative domain adaptation.
\newblock In {\em CVPR}, 2017.

\bibitem{DomainConfusion}
E. Tzeng, J. Hoffman, N. Zhang, K. Saenko, and T. Darrell.
\newblock Deep domain confusion: Maximizing for domain invariance.
\newblock {\em arXiv:1412.3474}.

\bibitem{tsne}
L.J.P. van~der Maaten and G.E. Hinton.
\newblock Visualizing high-dimensional data using t-sne.
\newblock {\em JMLR}, 9:2579--2605, 2008.

\bibitem{vicente2015driver}
F. Vicente, Z. Huang, X. Xiong, F. De~la Torre, W. Zhang, and D. Levi.
\newblock Driver gaze tracking and eyes off the road detection system.
\newblock {\em T-ITS}, 16(4):2014--2027, 2015.

\bibitem{vora2018driver}
S. Vora, A. Rangesh, and M. Trivedi.
\newblock Driver gaze zone estimation using convolutional neural networks: A
  general framework and ablative analysis.
\newblock {\em T-IV}, 3(3):254--265, 2018.

\bibitem{CosFace}
H. Wang et~al.
\newblock Cosface: Large margin cosine loss for deep face recognition.
\newblock In {\em CVPR}, 2018.

\bibitem{CenterLoss}
Y. Wen, K. Zhang, Z. Li, and Y. Qiao.
\newblock A discriminative feature learning approach for deep face recognition.
\newblock In {\em ECCV}, 2016.

\bibitem{wood2016learning}
E. Wood, T. Baltru{\v{s}}aitis, L.P. Morency, P. Robinson, and A. Bulling.
\newblock Learning an appearance-based gaze estimator from one million
  synthesised images.
\newblock In {\em ETRA}, 2016.

\bibitem{wood2015rendering}
E. Wood, T. Baltrusaitis, X. Zhang, Y. Sugano, P. Robinson, and A. Bulling.
\newblock Rendering of eyes for eye-shape registration and gaze estimation.
\newblock In {\em ICCV}, 2015.

\bibitem{yao2014gesture}
A. Yao, L. Van~Gool, and P. Kohli.
\newblock Gesture recognition portfolios for personalization.
\newblock In {\em CVPR}, 2014.

\bibitem{atrous}
F. Yu and V. Koltun.
\newblock Multi-scale context aggregation by dilated convolutions.
\newblock In {\em ICLR}, 2016.

\bibitem{yu2019improving}
Y. Yu, G. Liu, and J.M. Odobez.
\newblock Improving few-shot user-specific gaze adaptation via gaze redirection
  synthesis.
\newblock In {\em CVPR}, 2019.

\bibitem{zhang2017mpiigaze}
X. Zhang, Y. Sugano, M. Fritz, and A. Bulling.
\newblock Mpiigaze: Real-world dataset and deep appearance-based gaze
  estimation.
\newblock {\em T-PAMI}, 41(1):162--175, 2017.

\bibitem{RingLoss}
Y. Zheng, DK. Pal, and M. Savvides.
\newblock Ring loss: Convex feature normalization for face recognition.
\newblock In {\em CVPR}, 2018.

\bibitem{tri_train}
ZH. Zhou and M. Li.
\newblock Tri-training: Exploiting unlabled data using three classifiers.
\newblock {\em T-KDE.}, 17(11):1529--1541, 2005.

\end{thebibliography}
}

\end{document}